\documentclass[pdflatex,sn-mathphys-num]{sn-jnl}

\usepackage{graphicx}%
\usepackage{multirow}%
\usepackage{amsmath,amssymb,amsfonts}%
\usepackage{amsthm}%
\usepackage{mathrsfs}%
\usepackage[title]{appendix}%
\usepackage{xcolor}%
\usepackage{textcomp}%
\usepackage{manyfoot}%
\usepackage{booktabs}%
\usepackage{listings}%
\usepackage{bm,todonotes}
\usepackage{algorithm}%
\usepackage{algorithmic}%
\usepackage{listings}%

\theoremstyle{thmstyleone}%
\newtheorem{thm}{Theorem}
\newtheorem{lem}{Lemma}
\newtheorem{cor}{Corollary}
\newtheorem{prop}{Proposition}

\newtheorem{defn}{Definition}

\newtheorem{rem}{Remark}

\newtheorem{prf}{Proof}
\DeclareMathOperator{\Tr}{Tr}
\theoremstyle{thmstyletwo}%

\theoremstyle{thmstylethree}%

\begin{document}

\title[Enhancing Differentially Private Linear Regression via Public Second-Moment]{Enhancing Differentially Private Linear Regression via Public Second-Moment}


\author[1]{\fnm{Zilong} \sur{Cao}}\email{nwu\_czl@stumail.nwu.edu.cn}

\author*[2]{\fnm{Hai} \sur{Zhang}}\email{zhanghai@nwu.edu.cn}

\affil[1,2]{\orgdiv{The School of Mathematics}, \orgname{Northwest University}, \orgaddress{\city{Xi'an}, \postcode{710100}, \state{Shannxi}, \country{China}}}
\abstract{Leveraging information from public data has become increasingly crucial in enhancing the utility of differentially private (DP) methods. Traditional DP approaches often require adding noise based solely on private data, which can significantly degrade utility. In this paper, we address this limitation in the context of the ordinary least squares estimator (OLSE) of linear regression based on sufficient statistics perturbation (SSP) under the unbounded data assumption. We propose a novel method that involves transforming and clipping private data using the public second-moment matrix to compute a transformed SSP-OLSE. The second-moment matrix of transformed data yields a better condition number and improves the SSP-OLSE accuracy and robustness. We derive theoretical error bounds about our method and the standard SSP-OLSE to the non-DP OLSE, which reveal the improved robustness and accuracy achieved by our approach. Experiments on synthetic and real-world datasets demonstrate the utility and effectiveness of our method.}



\maketitle
\section{Introduction}\label{sec1}

Differential privacy (DP), introduced by \citet{dwork2014algorithmic}, provides a rigorous framework for limiting the disclosure risk of individual-level data through randomized algorithms whose outputs are insensitive to any single sample. It has become a fundamental tool in privacy-preserving statistics and machine learning \citep{wang2022differentially, guo2023privacy,wasserman2010statistical,dwork2015robust,kamath2019privately,jain2014near,abadi2016deep,blanco2022critical,10538298}. A central difficulty in DP methodology, however, is the tension between privacy protection and statistical utility: to ensure privacy, one must inject noise into statistics computed from private data, but this often leads to instability and substantial accuracy loss.

In many applications, one also has access to auxiliary \emph{public} data that are non-sensitive and can therefore be used without consuming privacy budget. This observation has motivated a growing literature on combining public and private information in DP estimation and learning. For example, \citet{ferrando2021combining} study statistical estimation with mixed public-private samples; \citet{bie2022private} show that a small public sample can improve private Gaussian estimation; and \citet{nasr2023effectively} develop a DP gradient method that benefits from public information. Related ideas have also appeared in private query release, synthetic data generation, and predictive modeling \citep{ji2013differential,nandi2020privately,bassily2020learning,liu2021leveraging}. Public data need not enter merely as additional samples, it also provides some statistics, such as a public second-moment matrix. The public second-moment contains the second-order structure that captures the geometry of the covariates, which is important for the linear regression.

For DP linear regression, existing methods are broadly divided into two classes. Optimization-based approaches, including gradient perturbation and objective perturbation \citep{cai2021cost,arora2022differentially,chaudhuri2011differentially,kifer2012private}, are well suited to iterative learning procedures. In contrast, sufficient statistics perturbation (SSP) methods \citep{wang2018revisiting} privatize the empirical second-moment matrix and cross-moment vector directly, yielding a closed-form estimator closely tied to the OLS sufficient statistics. While this strategy is especially natural under bounded covariates, many regression problems involve unbounded support, in which sensitivity control requires an additional data clipping step. In such settings, the utility of SSP depends even more strongly on the underlying second-order geometry. Subsequent work has refined this line in different directions: \citet{sheffet2017differentially} incorporates the Johnson--Lindenstrauss Transform (JLT) to preserve statistical structure under privacy constraints, while \citet{milionis2022differentially} study private regression with unbounded Gaussian covariates through private moment estimation and related bias analysis. Our method is most closely related to SSP-based private linear regression, but differs in a crucial aspect: rather than relying only on private-side perturbation and estimation, we use auxiliary public second-order information to precondition the private regression geometry before privatization.

This distinction matters because existing SSP-style methods still depend entirely on private data. As a result, two difficulties remain. First, the sufficient statistics may have high sensitivity, especially beyond uniformly bounded settings, forcing the mechanism to add substantial noise. Second, the resulting utility depends critically on the conditioning of the empirical second-moment matrix. Once this matrix is privatized, ill-conditioning can make the inverse unstable and amplify estimation error. These issues are particularly acute for ordinary least squares estimation, whose accuracy depends directly on the quality of the empirical second-moment matrix and its inverse. Since both the sensitivity of the sufficient statistics and the numerical stability of the resulting estimator are closely tied to the geometry of the empirical second-moment matrix, this suggests that public second-order information may offer a principled way to improve SSP-based regression before privatization.

We study ordinary least squares estimation within the SSP framework and ask whether public second-order information can improve the geometry of the private regression problem before privatization. We answer this affirmatively by proposing a public second-moment transformation that preconditions the private design matrix prior to perturbing the sufficient statistics. The transformation is designed to improve the conditioning of the private design and to make the transformed covariates closer to isotropic. This, in turn, yields a tighter clipping radius of transformed samples, reduces the sensitivity of privatized sufficient statistics, and stabilizes the inverse appearing in the DP estimator. In this sense, the role of public data in our framework is not generic sample augmentation, but geometric preconditioning. Our main contributions are as follows:
\begin{itemize}
	\item \textbf{A public-moment preconditioning framework for DP linear regression.} We propose a public second-moment transformation tailored to SSP-based private ordinary least squares. Rather than using public data as additional samples, the method uses public second-order geometry to precondition the private design before privatization.

	\item \textbf{Improved conditioning, stability, and utility guarantees.} We show that the proposed transformation makes the regression geometry more favorable for private sufficient-statistics perturbation, leading to better control of sensitivity of the transformed sufficient statistics, a more stable privatized second-moment inverse, and sharper non-asymptotic utility bounds than standard SSP-based private OLS under the benchmark regime.

	\item \textbf{Empirical evidence on synthetic and real data.} We conduct experiments on synthetic and real-world datasets and show that the proposed method consistently outperforms private-only baselines. The empirical findings support the theoretical message that an appropriate public second-moment preconditioner can substantially improve the stability and accuracy of DP linear regression.
\end{itemize}

The remainder of this paper is organized as follows. Section \ref{sec:statement} introduces the problem setup and presents the motivation. Section \ref{sec:Main Theoretical Results} establishes the main theoretical guarantees, including the stability condition and utility bounds. Section \ref{sec:experiment} reports the experimental results on synthetic and real datasets. Section \ref{sec:conclusion} concludes with a discussion of limitations and future directions.

\section{Problem Setup and Motivation}\label{sec:statement}

This section introduces the privacy framework and the regression setting used throughout the paper. We first review the privacy notions needed for sufficient statistics perturbation. We then formulate private linear regression with auxiliary public data, explain the sources of instability in standard private-only sufficient statistics perturbation, and motivate our public second-moment transformation. Technical tools used later in the theoretical analysis are deferred to Appendix~\ref{app:technical_tools} to keep the main development focused.

\subsection{Privacy background}\label{subsec:DP}

We use zero-concentrated differential privacy (zCDP) as the main privacy accounting framework, and convert it to $(\varepsilon,\delta)$-DP \citep{dwork2014algorithmic} when needed.

\begin{defn}[zCDP \citep{bun2016concentrated}]
Let $\mathcal{M}:\mathcal{X}^n \to \mathcal{S}$ be a randomized algorithm. We say that $\mathcal{M}$ satisfies $\rho$-zCDP if, for every pair of neighboring datasets $\mathbf{X},\mathbf{X}' \in \mathcal{X}^n$ and every $\alpha \in (1,\infty)$,
\[
D_{\alpha}\!\left(\mathcal{M}(\mathbf{X}) \,\|\, \mathcal{M}(\mathbf{X}')\right)
\le \rho \alpha,
\]
where $D_{\alpha}(\cdot\|\cdot)$ denotes the Rényi divergence of order $\alpha$. 
\end{defn}

We adopt zCDP because our method perturbs a matrix query and a vector query separately, and zCDP provides a convenient additive composition rule for the corresponding Gaussian mechanisms.

\begin{lem}[zCDP composition \citep{bun2016concentrated}]\label{lem:composition}
If $\mathcal{M}_1,\ldots,\mathcal{M}_T$ satisfy $\rho_1,\ldots,\rho_T$-zCDP, respectively, then the joint release $(\mathcal{M}_1,\ldots,\mathcal{M}_T)$ satisfies $\bigl(\sum_{t=1}^T \rho_t\bigr)$-zCDP.
\end{lem}

To calibrate Gaussian perturbations, we use the usual notion of $\ell_2$-sensitivity.

\begin{defn}[Sensitivity]
Let $f:\mathcal{X}^n \to \mathbb{R}^m$. Its $\ell_2$-sensitivity is
\[
\Delta_f
=
\sup_{\mathbf{X}\sim \mathbf{X}'}
\|f(\mathbf{X}) - f(\mathbf{X}')\|_2,
\]
where $\mathbf{X}\sim \mathbf{X}'$ means that the two datasets differ in one sample. For matrix-valued queries, the corresponding sensitivity is defined with respect to the Frobenius norm $\| \cdot\|_F$.
\end{defn}

\begin{lem}[Gaussian mechanism]
Let $f:\mathcal{X}^n \to \mathbb{R}^m$ have $\ell_2$-sensitivity $\Delta_f$. Then
\[
\mathcal{M}(\mathbf{X})
=
f(\mathbf{X})
+
\mathbf{Z},
\qquad
\mathbf{Z}\sim \mathcal{N}\!\left(0,\frac{\Delta_f^2}{2\rho}\mathbf{I}_m\right),
\]
satisfies $\rho$-zCDP.
\end{lem}
Here $\rho>0$ is the privacy budget selected by the user, with smaller $\rho$ corresponding to stronger privacy protection and hence larger Gaussian perturbation. In our algorithm, the public second-moment matrix is computed solely from public data and therefore does not consume privacy budget; privacy is used only when perturbing the transformed private sufficient statistics.

\subsection{Private linear regression and sufficient statistics perturbation}\label{subsec:DPLR}

We consider the linear regression model
\[
y_i = \mathbf{x}_i^\top \bm{\beta}^\star + \varepsilon_i,
\qquad i=1,\dots,n,
\]
where $\mathbf{x}_i\in\mathbb{R}^d$ is the covariate vector, $\bm{\beta}^\star\in\mathbb{R}^d$ is the target coefficient, and the noise variables $\varepsilon_i$ satisfy $\mathbb{E}(\varepsilon_i\mid \mathbf{x}_i)=0$.

Throughout the paper, we assume access to
\[
\mathbf{A}\in\mathbb{R}^{n_{\scriptscriptstyle \! A}\times d},
\qquad
\mathbf{y}_{\scriptscriptstyle \!\!A}\in\mathbb{R}^{n_{\scriptscriptstyle \! A}},
\]
as the \emph{private} covariate matrix and response vector, and
\[
\mathbf{B}\in\mathbb{R}^{n_{\scriptscriptstyle \! B}\times d},
\qquad
\mathbf{y}_{\scriptscriptstyle \!\!B}\in\mathbb{R}^{n_{\scriptscriptstyle \! B}},
\]
as the \emph{public} covariate matrix and response vector. The private data are protected by differential privacy; the public data can be used freely. In the benchmark regime studied here, the rows of $\mathbf A$ and $\mathbf B$ are independently sampled from sub-Gaussian distributions with a common second-moment
\[
\Sigma = \mathbb{E}(\mathbf{x}\mathbf{x}^\top).
\]

For the private sample, define the empirical second-moment and cross-moment by
\[
\hat{\Sigma}_{\scriptscriptstyle \!A}
=
\frac{1}{n_{\scriptscriptstyle \! A}}\mathbf{A}^\top \mathbf{A},
\qquad
\hat{\Sigma}_{\scriptscriptstyle Ay}
=
\frac{1}{n_{\scriptscriptstyle \! A}}\mathbf{A}^\top \mathbf{y}_{\scriptscriptstyle \!\!A}.
\]
Whenever $\hat{\Sigma}_{\scriptscriptstyle A}$ is invertible, the private ordinary least squares estimator is
\[
\hat{\bm{\beta}}_{\scriptscriptstyle A}
=
\hat{\Sigma}_{\scriptscriptstyle A}^{-1}\hat{\Sigma}_{\scriptscriptstyle Ay}.
\]

A standard private approach is \emph{sufficient statistics perturbation} (SSP), which privatizes $\hat{\Sigma}_{\scriptscriptstyle A}$ and $\hat{\Sigma}_{\scriptscriptstyle Ay}$ directly:
\[
\hat{\bm{\beta}}^{\,DP}_{\scriptscriptstyle A}
=
\bigl(\hat{\Sigma}_{\scriptscriptstyle A}+\mathbf{G}\bigr)^{-1}
\bigl(\hat{\Sigma}_{\scriptscriptstyle Ay}+\mathbf{g}\bigr),
\]
where $\mathbf{G}$ and $\mathbf{g}$ are Gaussian perturbations calibrated to the sensitivities of the released sufficient statistics.

This approach is attractive because it is closed-form and directly targets the OLS sufficient statistics, but its performance is governed by the scale and conditioning of the empirical second-moment. In the private-only setting it faces two structural difficulties.

\textbf{High sensitivity under unbounded or anisotropic covariates.} When response and covariates are unbounded, SSP typically requires data clipping to control sensitivities of $\hat{\Sigma}_{\scriptscriptstyle A}$ and $\hat{\Sigma}_{\scriptscriptstyle Ay}$. However, when the covariates are highly anisotropic, clip in the original coordinates may be poorly aligned with the covariance geometry, which can either induce excessive distortion or lead to unnecessarily large DP noise.

\textbf{Instability of the privatized inverse.} Even without privacy, OLS is sensitive to the conditioning of $\hat{\Sigma}_{\scriptscriptstyle A}$. After adding Gaussian perturbation, the matrix $(\hat{\Sigma}_{\scriptscriptstyle A}+\mathbf{G})^{-1}$ may become substantially less stable and amplify the perturbation error. This effect is particularly severe when $\hat{\Sigma}_{\scriptscriptstyle A}$ is ill-conditioned.

These two issues are both geometric: they depend on the scale and conditioning of the empirical second-moment matrix. This motivates using public second-order information to improve the geometry \emph{before} privatization.

\subsection{Public second-moment transformation and regression}\label{subsec:motivation}

Let
\[
\hat{\Sigma}_{\scriptscriptstyle B}
=
\frac{1}{n_{\scriptscriptstyle \! B}}\mathbf{B}^\top \mathbf{B}
\]
be the public empirical second-moment matrix. Since $\mathbf{B}$ is public, $\hat{\Sigma}_{\scriptscriptstyle B}$ can be computed without consuming privacy budget.

Our basic idea is to use $\hat{\Sigma}_{\scriptscriptstyle B}$ as a preconditioner for the private covariates. Define the transformed private design matrix by
\[
\tilde{\mathbf{A}}
=
\mathbf{A}\hat{\Sigma}_{\scriptscriptstyle B}^{-1/2},
\]
then the transformed covariates are closer to isotropic:
\[
\mathbb{E}\!\left(\tilde{\mathbf{x}}\tilde{\mathbf{x}}^\top\right)
=
\hat{\Sigma}_{\scriptscriptstyle B}^{-1/2}\Sigma \hat{\Sigma}_{\scriptscriptstyle B}^{-1/2}
\approx
\mathbf{I}_d.
\]
This has two consequences that are central for private regression.

First, the transformed covariates admit more homogeneous norm control, so the clipping can be performed in a geometry aligned with the covariance structure rather than in the original anisotropic coordinates. This improves the sensitivity bounds of the privatized sufficient statistics.

Second, the transformed empirical second-moment
\[
\tilde{\Sigma}_{\scriptscriptstyle A}
=
\frac{1}{n_{\scriptscriptstyle \! A}}\tilde{\mathbf{A}}^\top \tilde{\mathbf{A}}
=
\hat{\Sigma}_{\scriptscriptstyle B}^{-1/2}\hat{\Sigma}_{\scriptscriptstyle A}\hat{\Sigma}_{\scriptscriptstyle B}^{-1/2}\approx
\mathbf{I}_d
\]
is typically better conditioned than $\hat{\Sigma}_{\scriptscriptstyle A}$. Since SSP ultimately requires inverting a perturbed second-moment matrix, improving conditioning before perturbation directly enhances numerical stability and downstream utility.

For the response variable, we rescale by the public response second-moment
\[
\hat{\sigma}_{\scriptscriptstyle \!y,B}
=
\left(\frac{1}{n_{\scriptscriptstyle \! B}}\|\mathbf{y}_{\scriptscriptstyle \!\!B}\|_2^2\right)^{1/2},
\qquad
\tilde{\mathbf{y}}_{\scriptscriptstyle \!\!A}
=
\frac{\mathbf{y}_{\scriptscriptstyle \!\!A}}{\hat{\sigma}_{\scriptscriptstyle \!y,B}},
\]
and define
\[
\tilde{\Sigma}_{\scriptscriptstyle Ay}
=
\frac{1}{n_{\scriptscriptstyle \! A}}\tilde{\mathbf{A}}^\top \tilde{\mathbf{y}}_{\scriptscriptstyle \!\!A}.
\]
Although this scalar transformation is not the main source of improvement for the regression, it is introduced mainly to place the transformed cross-moment on a principled scale for the data clipping and sensitivity calibration.

For the parameter estimation, the transformation corresponds to a reparameterization plus a scalar rescaling, and therefore preserves the underlying OLS solution after mapping back. Indeed, define the transformed linear regression estimator
\begin{equation}\label{eq:transform}
	\begin{aligned}
		\bm{\tilde{\beta}}_{\scriptscriptstyle \!A} := (\tilde{\Sigma}_{\scriptscriptstyle \!A})^{-1}\tilde{\Sigma}_{\scriptscriptstyle \!Ay}
		&= \Big(\frac{\tilde{\mathbf{A}}^{\scriptscriptstyle \!\!\top} \tilde{\mathbf{A}}}{n_{\scriptscriptstyle \!A}}\Big)^{-1}\Big(\frac{\tilde{\mathbf{A}}^{\scriptscriptstyle \!\!\top} \tilde{\mathbf{y}}_{\scriptscriptstyle \!\!A}}{n_{\scriptscriptstyle \!A}}\Big) \\
		&= \Big(\frac{\hat{\Sigma}_{\scriptscriptstyle \!B}^{\scriptscriptstyle \!-\!1\!/\!2}\mathbf{A}^{\scriptscriptstyle \!\!\top} \!\mathbf{A}\hat{\Sigma}_{\scriptscriptstyle \!B}^{\scriptscriptstyle \!-\!1\!/\!2}}{n_{\scriptscriptstyle \!A}}\Big)^{-1} \Big(\frac{\hat{\Sigma}_{\scriptscriptstyle \!B}^{\scriptscriptstyle \!-\!1\!/\!2}\mathbf{A}^{\scriptscriptstyle \!\!\top}\mathbf{y}_{\scriptscriptstyle \!\!A} \hat{\sigma}_{\scriptscriptstyle \!y,B}^{\scriptscriptstyle -1}}{n_{\scriptscriptstyle \!A}} \Big)\\
		&= \hat{\sigma}_{\scriptscriptstyle \!y,B}^{\scriptscriptstyle -1}\hat{\Sigma}_{\scriptscriptstyle \!B}^{1/2}\bm{\hat{\beta}}_{\scriptscriptstyle \!A}. 	
	\end{aligned}
\end{equation}
Thus, the OLS estimator in the transformed domain is exactly the transformed version of the original OLS estimator. Moreover, the transformed second-moment $\tilde\Sigma_{\scriptscriptstyle A}$ is typically better conditioned and hence more robust to additive perturbation. Hence, Equation~\eqref{eq:transform} formalizes the key principle of our method: we may carry out clip, sufficient statistics perturbation, and stability analysis in the transformed domain, where the geometry is more favorable, and then recover the estimator in the original coordinates by post-processing. Namely, after computing a private estimator $\tilde{\bm{\beta}}_{\scriptscriptstyle A}^{\,DP}$ in the transformed domain, we map it back to the original domain by 
\[
\bar{\bm{\beta}}_{\scriptscriptstyle A}^{\,DP}
=
\hat{\sigma}_{\scriptscriptstyle \!y,B}\hat{\Sigma}_{\scriptscriptstyle B}^{\scriptscriptstyle - 1/2}\tilde{\bm{\beta}}_{\scriptscriptstyle A}^{\,DP},
\] which underlies our strategy for obtaining a more stable DP estimator after post-processing.

\begin{figure}[H]
  \centering
	\includegraphics[scale = 0.51]{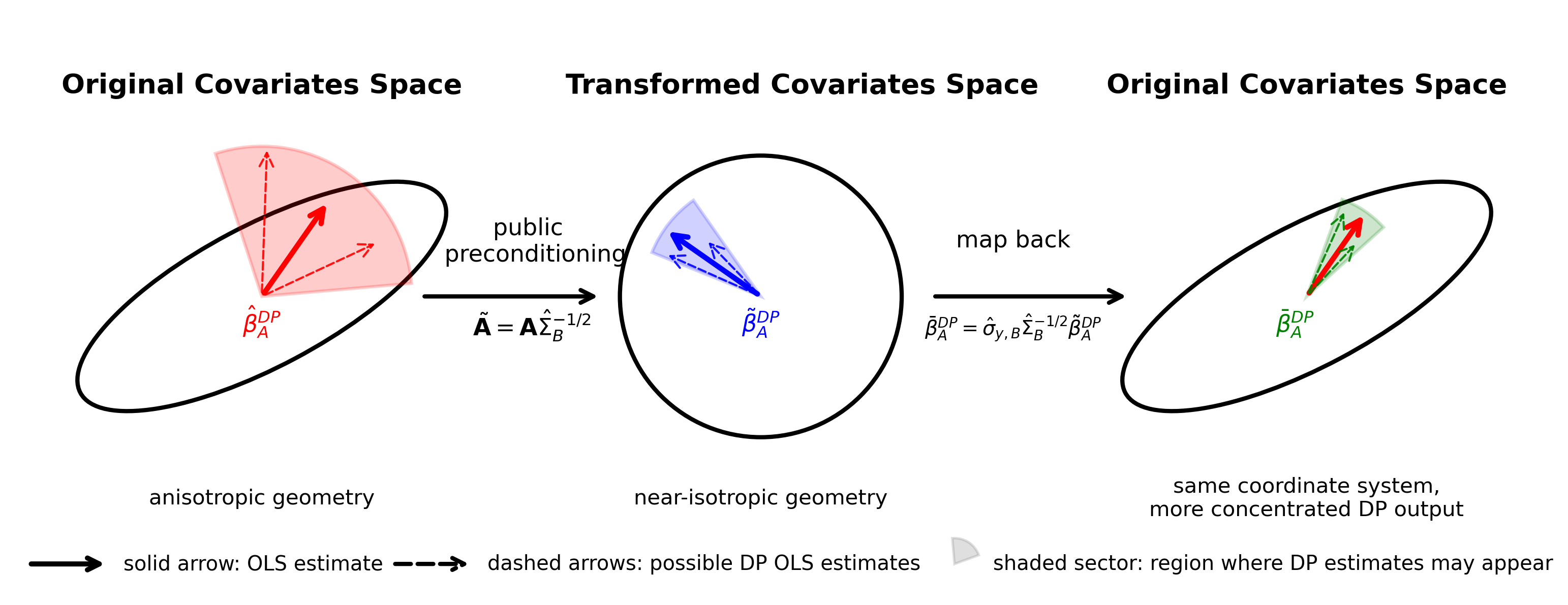}
	\caption{Schematic illustration of the estimator transformation. The original OLS estimator is first reparameterized in the public-moment-transformed domain, privatization is then performed on the transformed sufficient statistics, and the resulting DP estimator is finally mapped back to the original coordinates by post-processing.}\label{fig:estimator_transformation}
\end{figure}

\section{Main Method and Theoretical Guarantees}\label{sec:Main Theoretical Results}

Building on Section~\ref{sec:statement}, we now formalize the transformed-domain private regression procedure and establish its main guarantees. The key idea is to carry out clipping, sufficient-statistics perturbation, and inversion in the public-moment-preconditioned domain, where the covariates are closer to isotropic and the second-moment matrix is better conditioned, and then map the estimator back to the original coordinates by post-processing.
\begin{figure}[H]
  \centering
	\includegraphics[scale = 0.51]{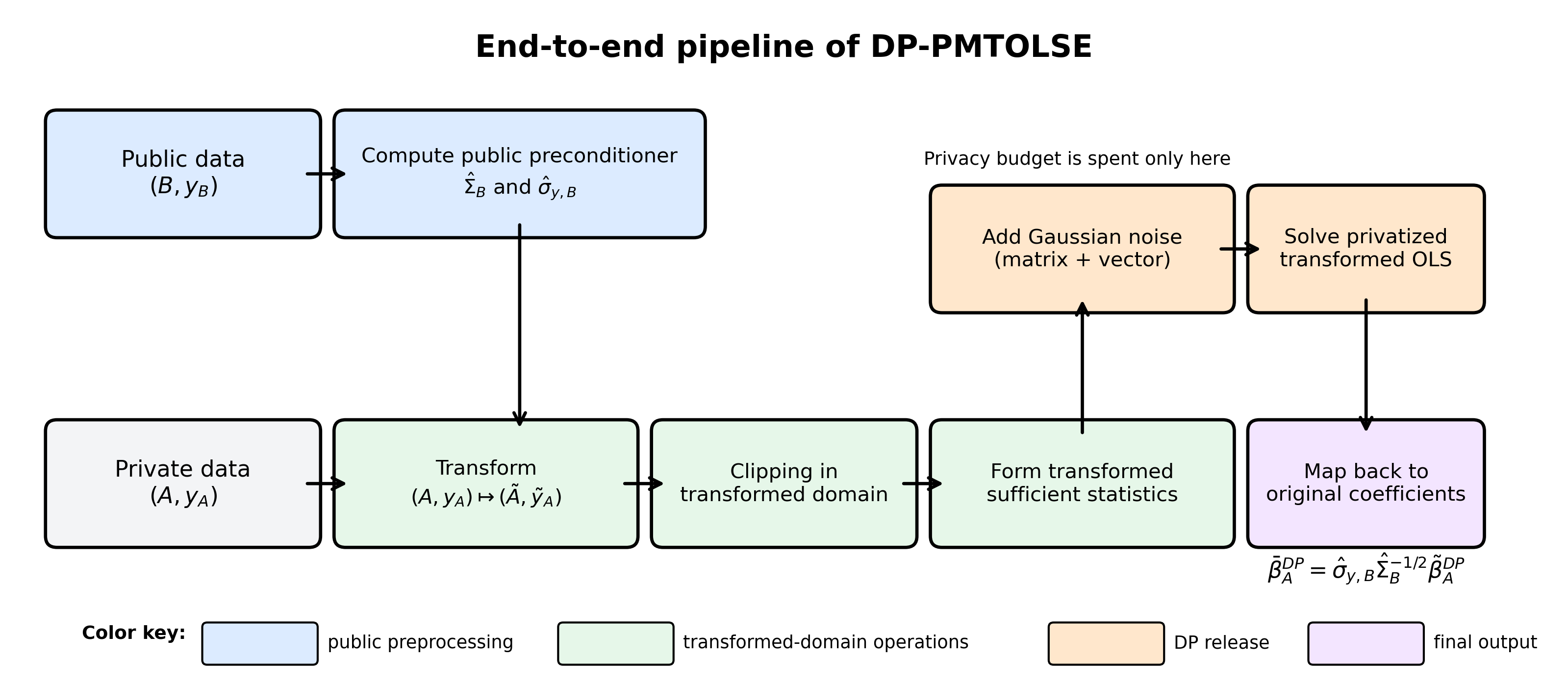}
	\caption{Pipeline of \textbf{Algorithm \ref{alg:DPPMTOLSE}}, DP-PMTOLSE.}\label{fig:pipeline}
\end{figure}

\subsection{Transformed-domain private regression procedure}\label{subsec:algorithm_main}

We first describe the linear regression procedure induced by the public second-moment transformation. Let
\[
r_{A}(n_{\scriptscriptstyle \!A},d,\eta)
=
\sqrt{d\!\left(1+\log\!\frac{2n_{\scriptscriptstyle \!A}}{\eta}\right)}
\]
be the clipping radius for transformed covariates, and let
\[
r_{y}(n_{\scriptscriptstyle \!A},\eta)
=
\sqrt{1+\log\!\frac{2n_{\scriptscriptstyle \!A}}{\eta}}
\]
be the corresponding radius for transformed responses after normalization. These radiuses are chosen according to the sub-Gaussian tail bounds in the approximately isotropic transformed domain.

For convenience, we split the total privacy budget into two parts:
\[
\rho_1+\rho_2=\rho_{\mathrm{tot}},
\]
where $\rho_1$ is used for the transformed second-moment matrix and $\rho_2$ is used for the transformed cross-moment vector. In the statements below, one may take the equal split $\rho_1=\rho_2=\rho_{\mathrm{tot}}/2$. Figure~\ref{fig:pipeline} summarizes the full procedure about \textbf{Algorithm~\ref{alg:DPPMTOLSE}}.

\begin{algorithm}[H]
	\small
	\caption{DP-PMTOLSE: Differentially Private OLS via Public Second-Moment Transformation}\label{alg:DPPMTOLSE}
	\begin{algorithmic}[1]
		\STATE {\bfseries Input:} private data $(\mathbf{A},\mathbf{y}_{\scriptscriptstyle \!\!A})$, public data $(\mathbf{B},\mathbf{y}_{\scriptscriptstyle \!\!B})$, privacy budget $\rho_{\mathrm{tot}}$, failure probability $\eta$.
		
		\STATE {\bfseries Compute public preconditioner:}
		\[
		\hat{\Sigma}_{\scriptscriptstyle \!B}
		=
		\frac{1}{n_{\scriptscriptstyle \!B}}\mathbf{B}^{\top}\mathbf{B},
		\qquad
		\hat{\sigma}_{\scriptscriptstyle \!y,B}
		=
		\left(\frac{1}{n_{\scriptscriptstyle \!B}}\|\mathbf{y}_{\scriptscriptstyle \!\!B}\|_2^2\right)^{1/2}.
		\]
		
		\STATE {\bfseries Transform private covariates and responses:}
		\[
		\tilde{\mathbf{A}}
		=
		\mathbf{A}\hat{\Sigma}_{\scriptscriptstyle \!B}^{-1/2},
		\qquad
		\tilde{\mathbf{y}}_{\scriptscriptstyle \!\!A}
		=
		\mathbf{y}_{\scriptscriptstyle \!\!A}/\hat{\sigma}_{\scriptscriptstyle \!y,B}.
		\]
		
		\STATE {\bfseries Clipping transformed covariates:} for each row $\tilde{\mathbf{a}}_i$ of $\tilde{\mathbf{A}}$,
		\[
		\tilde{\mathbf{a}}_i
		\leftarrow
		\tilde{\mathbf{a}}_i\cdot
		\min\!\left\{
		1,\,
		\frac{r_A(n_{\scriptscriptstyle \!A},d,\eta)}{\|\tilde{\mathbf{a}}_i\|_2}
		\right\}.
		\]
		
		\STATE {\bfseries Clipping transformed responses:} for each entry $\tilde y_{\scriptscriptstyle \!A,i}$ of $\tilde{\mathbf{y}}_{\scriptscriptstyle \!\!A}$,
		\[
		\tilde y_{\scriptscriptstyle \!A,i}
		\leftarrow
		\tilde y_{\scriptscriptstyle \!A,i}\cdot
		\min\!\left\{
		1,\,
		\frac{r_y(n_{\scriptscriptstyle \!A},\eta)}{|\tilde y_{\scriptscriptstyle \!A,i}|}
		\right\}.
		\]
		
		\STATE {\bfseries Form transformed sufficient statistics:}
		\[
		\tilde{\Sigma}_{\scriptscriptstyle A}
		=
		\frac{1}{n_{\scriptscriptstyle \!A}}\tilde{\mathbf{A}}^\top\tilde{\mathbf{A}},
		\qquad
		\tilde{\Sigma}_{\scriptscriptstyle Ay}
		=
		\frac{1}{n_{\scriptscriptstyle \!A}}\tilde{\mathbf{A}}^\top\tilde{\mathbf{y}}_{\scriptscriptstyle \!\!A}.
		\]
		
		\STATE {\bfseries Add Gaussian noise:}
		\[
		\tilde{\Sigma}_{\scriptscriptstyle A}^{\,DP}
		=
		\tilde{\Sigma}_{\scriptscriptstyle A}
		+\mathbf{G},
		\qquad
		\tilde{\Sigma}_{\scriptscriptstyle Ay}^{\,DP}
		=
		\tilde{\Sigma}_{\scriptscriptstyle Ay}
		+\mathbf{g},
		\]
		where
		\[
		\mathbf{G}\sim GUE(\sigma_1^2),
		\qquad
		\mathbf{g}\sim \mathcal N(0,\sigma_2^2\mathbf{I}_d),
		\]
		and $\sigma_1,\sigma_2$ are calibrated from the sensitivities in Proposition~\ref{prop:sensitivity_privacy}.
		
		\STATE {\bfseries Solve the privatized transformed regression:}
		\[
		\tilde{\bm{\beta}}_{\scriptscriptstyle A}^{\,DP}
		=
		\bigl(\tilde{\Sigma}_{\scriptscriptstyle A}^{\,DP}\bigr)^{-1}
		\tilde{\Sigma}_{\scriptscriptstyle Ay}^{\,DP}.
		\]
		
		\STATE {\bfseries Map back to the original coordinates:}
		\[
		\bar{\bm{\beta}}_{\scriptscriptstyle A}^{\,DP}
		=
		\hat{\sigma}_{\scriptscriptstyle \!y,B}
		\hat{\Sigma}_{\scriptscriptstyle \!B}^{-1/2}
		\tilde{\bm{\beta}}_{\scriptscriptstyle A}^{\,DP}.
		\]
		
		\STATE {\bfseries Output:} $\bar{\bm{\beta}}_{\scriptscriptstyle A}^{\,DP}$.
	\end{algorithmic}
\end{algorithm}

\textbf{Algorithm~\ref{alg:DPPMTOLSE}} improves the private regression problem along two coupled directions. The public second-moment transformation replaces anisotropic clip in the original coordinates by approximately isotropic clip in the transformed coordinates, and it makes the privatized second-moment inverse less sensitive to Gaussian perturbation. The next subsection formalizes these two effects.

\subsection{Spectral regularization and transformed-domain clip}\label{subsec:spectral_clip}

\begin{figure}[H]
  \centering
	\includegraphics[scale = 0.51]{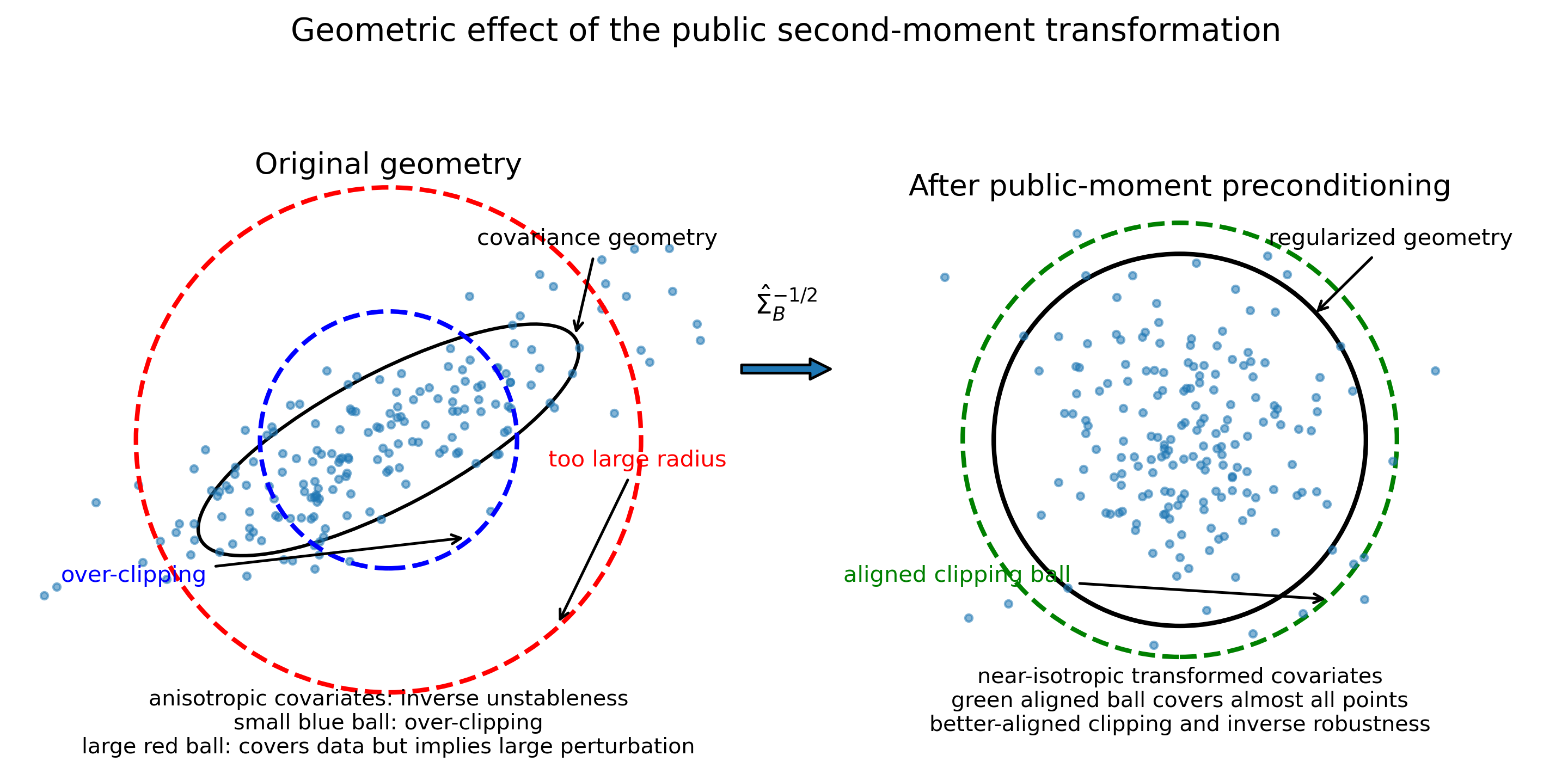}
	\caption{Schematic illustration of the data clipping. The public preconditioning turns anisotropic covariates into approximately isotropic ones, enabling dimension-scale clip.}\label{fig:clip}
\end{figure}

We first show that the public second-moment transformation regularizes the transformed covariance structure. This is the core geometric reason why the transformed regression problem is easier to privatize. Figure~\ref{fig:clip} provides a schematic illustration of the geometric effect of preconditioning on data clipping.

\begin{lem}[Spectral regularization induced by public preconditioning]\label{lem:precondition_spectrum}
Let $\mathbf{a}\in\mathbb{R}^d$ be a sub-Gaussian random vector with second-moment $\Sigma=\mathbb{E}(\mathbf{a}\mathbf{a}^\top)$,
and let $\mathbf{B}\in\mathbb{R}^{n_{\scriptscriptstyle \!B}\times d}$ have i.i.d.\ rows distributed as $\mathbf{a}$. Define
\[
\hat{\Sigma}_{\scriptscriptstyle \!B}
=
\frac{1}{n_{\scriptscriptstyle \!B}}\mathbf{B}^\top\mathbf{B}.
\]
Then the transformed vector
\[
\tilde{\mathbf{a}}
=
\hat{\Sigma}_{\scriptscriptstyle \!B}^{-1/2}\mathbf{a}
\]
is sub-Gaussian with second-moment
\[
\tilde{\Sigma}
=
\hat{\Sigma}_{\scriptscriptstyle \!B}^{-1/2}
\Sigma
\hat{\Sigma}_{\scriptscriptstyle \!B}^{-1/2}.
\]
Moreover, for every $\eta\in(0,1)$, with probability at least $1-2\eta$,
\[
L\,\mathbf{I}_d
\preceq
\tilde{\Sigma}
\preceq
U\,\mathbf{I}_d,
\]
where
\[
L= \Big(1 + C\Big(\sqrt{\frac{d}{n_{\scriptscriptstyle \!B}}} + \sqrt{\frac{\log(1/\eta)}{n_{\scriptscriptstyle \!B}}}\Big)\Big)^{-2},
\qquad
U = \Big(1 - C\Big(\sqrt{\frac{d}{n_{\scriptscriptstyle \!B}}} + \sqrt{\frac{\log(1/\eta)}{n_{\scriptscriptstyle \!B}}}\Big)\Big)^{-2},
\]

for the constant $C$ depending only on the sub-Gaussian norm.
\end{lem}

\textbf{Lemma~\ref{lem:precondition_spectrum}} shows that, under the matched second-moment benchmark regime, the transformed covariates are approximately isotropic whenever the public sample is large enough, e.g., a little larger than the dimension $n_{\scriptscriptstyle \!B} \gtrsim  d + \log(1/\eta)$. Moreover, the transformed covariance has eigenvalues bounded away from $0$ and $\infty$ by quantities tending to $1$ fast as $n_{\scriptscriptstyle \!B}$ grows. In particular, $U$ provides the bounded clipping radius and $L$ bound the norm of the inverse second-moment matrix. They are the most important indicators to control the robustness of the whole problem. 

This yields a dimension-scale clipping radius independent of the unknown private covariance geometry. The following corollary is from \textbf{Lemma \ref{lem:nonisotropic_norm}} in Appendix \ref{app:technical_tools}.

\begin{cor}[Dimension-scale clip after public preconditioning]\label{cor:dimension_clip}
Under the assumptions of \textbf{Lemma~\ref{lem:precondition_spectrum}}, let $\tilde{\mathbf{a}}_1,\ldots,\tilde{\mathbf{a}}_{n_{\scriptscriptstyle \!A}}$ be i.i.d.\ transformed private covariates. Then for every $\eta\in(0,1)$, with probability at least $1-2\eta$,
\[
\|\tilde{\mathbf{a}}_i\|_2^2
\le
\Tr(\tilde{\Sigma})
+
C' d\log\!\left(\frac{2n_{\scriptscriptstyle \!A}}{\eta}\right)
\le
C'' d\!\left(1+\log\!\frac{2n_{\scriptscriptstyle \!A}}{\eta}\right),
\qquad
i=1,\ldots,n_{\scriptscriptstyle \!A},
\]
for constants $C',C''>0$ depending only on the sub-Gaussian norm. Here, $\Tr(\tilde{\Sigma}) \simeq d\big(\frac{L+U}{2}\big) \simeq  d$ with the sufficient large $n_{\scriptscriptstyle B}$. 
\end{cor}

\begin{rem}
\textbf{Corollary~\ref{cor:dimension_clip}} is the transformed-domain analogue of isotropic clip. The important point is not merely that the clip is possible, but that its radius depends only on dimension, the number of private samples, and failure probability up to constants, rather than on the unknown anisotropy of the private covariates. 
\end{rem}

\subsection{Sensitivity calibration and privacy guarantee}\label{subsec:sensitivity_privacy}

We next calibrate the Gaussian perturbations for the transformed sufficient statistics. Since the public transformation itself is computed only from public data, it does not consume privacy budget. Privacy is used only when releasing the transformed private second-moment and transformed private cross-moment.

\begin{prop}[Sensitivity and privacy guarantee for transformed sufficient statistics]\label{prop:sensitivity_privacy}
Consider Algorithm~\ref{alg:DPPMTOLSE}. The transformed sufficient statistics satisfy the sensitivity bounds
\[
\Delta_{\tilde{\Sigma}_{\scriptscriptstyle A}}
\le
\frac{2d\left(1+\log\!\frac{2n_{\scriptscriptstyle \!A}}{\eta}\right)}
{n_{\scriptscriptstyle \!A}},
\]
with respect to the Frobenius norm, and
\[
\Delta_{\tilde{\Sigma}_{\scriptscriptstyle Ay}}
\le
\frac{2\sqrt d\left(1+\log\!\frac{2n_{\scriptscriptstyle \!A}}{\eta}\right)}
{n_{\scriptscriptstyle \!A}},
\]
with respect to the Euclidean norm.

Therefore, if
\[
\sigma_1
=
\frac{\Delta_{\tilde{\Sigma}_{\scriptscriptstyle A}}}{\sqrt{2\rho_1}},
\qquad
\sigma_2
=
\frac{\Delta_{\tilde{\Sigma}_{\scriptscriptstyle Ay}}}{\sqrt{2\rho_2}},
\]
then releasing
\[
\tilde{\Sigma}_{\scriptscriptstyle A}^{\,DP}
=
\tilde{\Sigma}_{\scriptscriptstyle A}+\mathbf{G},
\qquad
\tilde{\Sigma}_{\scriptscriptstyle Ay}^{\,DP}
=
\tilde{\Sigma}_{\scriptscriptstyle Ay}+\mathbf{g},
\]
with
\[
\mathbf{G}\sim GUE(\sigma_1^2),
\qquad
\mathbf{g}\sim \mathcal N(0,\sigma_2^2\mathbf{I}_d),
\]
satisfies $(\rho_1+\rho_2)$-zCDP.
\end{prop}

\begin{prf}
The sensitivity bounds follow directly from the clipping radius:
\[\Delta_{\tilde{\Sigma}_{\scriptscriptstyle A}} =  \max\limits_{\tilde{\mathbf{a}},\tilde{\mathbf{a}}'}
\left\|
\frac{1}{n_{\scriptscriptstyle \!A}}
\bigl(\tilde{\mathbf{a}}\tilde{\mathbf{a}}^\top
-
\tilde{\mathbf{a}}'\tilde{\mathbf{a}}'^\top\bigr)
\right\|_F
\le
\frac{\|\tilde{\mathbf{a}}\|_2^2+\|\tilde{\mathbf{a}}'\|_2^2}
{n_{\scriptscriptstyle \!A}},
\]
and
\[\Delta_{\tilde{\Sigma}_{\scriptscriptstyle Ay}} = \max\limits_{\tilde{\mathbf{a}},\tilde{\mathbf{a}}',\tilde{y},\tilde{y}'}
\left\|
\frac{1}{n_{\scriptscriptstyle \!A}}
\bigl(\tilde{\mathbf{a}}\tilde y
-
\tilde{\mathbf{a}}'\tilde y'\bigr)
\right\|_2
\le
\frac{\|\tilde{\mathbf{a}}\|_2|\tilde y|+\|\tilde{\mathbf{a}}'\|_2|\tilde y'|}
{n_{\scriptscriptstyle \!A}}.
\] From the clip, we have 
\[
\|\tilde{\mathbf{a}}\|_2 \leq \sqrt{d\big(1 + \log\frac{2n_{\scriptscriptstyle \!A}}{\eta} \big)},\ \ \ |\tilde{y}| \leq \sqrt{(1 + \log\frac{2n_{\scriptscriptstyle \!A}}{\eta} \big)}.
\]

From the Gaussian mechanism, $\tilde{\Sigma}_{\scriptscriptstyle A}^{\,DP}$ satisfies $\rho_1$-zCDP and $\tilde{\Sigma}_{\scriptscriptstyle Ay}^{\,DP}$ satisfies $\rho_2$-zCDP. The additive composition rule is stated in \textbf{Lemma~\ref{lem:composition}}.
\end{prf}

To connect the privacy calibration with later utility bounds, we record high-probability bounds for the Gaussian perturbations.

\begin{cor}[High-probability bounds for privacy noise]\label{cor:noise_bounds}
Under the setting of Proposition~\ref{prop:sensitivity_privacy}, for every $\eta\in(0,1)$, with probability at least $1-2\eta$,
\[
\|\mathbf{G}\|_2
\le
C_1
\frac{d^{3/2}\left(1+\log\!\frac{2n_{\scriptscriptstyle \!A}}{\eta}\right)}
{\sqrt{\rho_1}\,n_{\scriptscriptstyle \!A}},
\]
and
\[
\|\mathbf{g}\|_2
\le
C_2
\frac{d\left(1+\log\!\frac{2n_{\scriptscriptstyle \!A}}{\eta}\right)}
{\sqrt{\rho_2}\,n_{\scriptscriptstyle \!A}},
\]
for constants $C_1,C_2>0$.
\end{cor}

\begin{rem}
The proof is the direct conclusion from \textbf{Lemma \ref{lem:GUEtail}} and \textbf{Lemma \ref{lem:gaussiannorm}} in Appendix \ref{app:Gaussian_tail}. The matrix perturbation is of order
\[
\|\mathbf{G}\|_2
=
O_{\mathbb P}\!\left(
\frac{d^{3/2}(1+\log(2n_{\scriptscriptstyle \!A}/\eta))}
{\sqrt{\rho_1}\,n_{\scriptscriptstyle \!A}}
\right),
\]
which is the key quantity entering the inverse-stability condition.
\end{rem}

\subsection{Main utility guarantee}\label{subsec:main_utility}

We now state the main guarantee for \textbf{Algorithm~\ref{alg:DPPMTOLSE}}. The theorem has three parts. First, it records the privacy guarantee. Second, it shows that no sample is clipped with high probability. Third, it bounds the error of the transformed-domain DP estimator and of the final estimator after mapping back to the original coordinates.

\begin{thm}[Privacy and utility of DP-PMTOLSE]\label{thm:DPPMTOLSE_main}
Suppose the private and public covariates are sub-Gaussian and satisfy the matched second-moment benchmark regime described in Section~\ref{subsec:DPLR}. Let $\bar{\bm{\beta}}_{\scriptscriptstyle A}^{\,DP}$ be the output of Algorithm~\ref{alg:DPPMTOLSE}, and define the transformed non-private OLS estimator by
\[
\tilde{\bm{\beta}}_{\scriptscriptstyle A}
=
\tilde{\Sigma}_{\scriptscriptstyle A}^{-1}\tilde{\Sigma}_{\scriptscriptstyle Ay}
=
\hat{\sigma}_{\scriptscriptstyle y,B}^{-1}
\hat{\Sigma}_{\scriptscriptstyle B}^{1/2}
\hat{\bm{\beta}}_{\scriptscriptstyle A},
\]
where $\tilde{\bm{\beta}}_{\scriptscriptstyle A}$ is the transformed-domain OLS estimator and $\hat{\bm{\beta}}_{\scriptscriptstyle A}$ is the original-domain OLS estimator.

For $\eta\in(0,1)$, write
\[
\Gamma_{A,\eta}
:=
\left(1+\log\!\frac{2n_{\scriptscriptstyle A}}{\eta}\right),
\qquad
\delta_{A,\eta}
:=
C\!\left(
\sqrt{\frac{d}{n_{\scriptscriptstyle A}}}
+
\sqrt{\frac{\log(1/\eta)}{n_{\scriptscriptstyle A}}}
\right),
\]
and
\[
L
:=
\frac{n_{\scriptscriptstyle B}}
{\left(
\sqrt{n_{\scriptscriptstyle B}}
+
C'\left(\sqrt d
+
\sqrt{\log\frac{1}{\eta}}\right)
\right)^2},
\]
where $C'>0$ is a constant depending only on the sub-Gaussian norms.

Then the following statements hold.

\begin{enumerate}
    \item[(i)] \textbf{Privacy.}
    Algorithm~\ref{alg:DPPMTOLSE} satisfies $\rho_{\mathrm{tot}}$-zCDP. In particular, one may take the equal budget split
    \[
    \rho_1=\rho_2=\frac{\rho_{\mathrm{tot}}}{2}.
    \]

    \item[(ii)] \textbf{Preservation under clip.}
    There exists a constant $c_0>0$ such that, with probability at least $1-c_0\eta$,
    \[
    \max_{1\le i\le n_{\scriptscriptstyle A}}\|\tilde{\mathbf A}_i\|_2 \le r_A(n_{\scriptscriptstyle A},d,\eta),
    \qquad
    \max_{1\le i\le n_{\scriptscriptstyle A}}|\tilde y_{\scriptscriptstyle A,i}| \le r_y(n_{\scriptscriptstyle A},\eta).
    \]
    Consequently, on this event the clip step leaves the transformed sufficient statistics unchanged, and hence preserves the transformed OLS target $\tilde{\bm{\beta}}_{\scriptscriptstyle A}$.

    \item[(iii)] \textbf{Utility under inverse stability.}
    Assume that $n_{\scriptscriptstyle B}$ is large enough so that $\hat{\Sigma}_{\scriptscriptstyle B}$ is invertible, which in the full-rank sub-Gaussian setting is ensured whenever
    \[
    n_{\scriptscriptstyle B}\gtrsim d+\log(1/\eta),
    \]
    and that the stability condition
    \begin{equation}\label{eq:stability_condition_main}
    \frac{d^{3/2}}{\sqrt{\rho_1}\,n_{\scriptscriptstyle A}\,}\frac{
    \Gamma_{A,\eta}
    }{ L(1-\delta_{A,\eta})^2
    }
    \le \frac{1}{2}
    \end{equation}
    holds.

    Then there exists a constant $C_0>0$, depending only on the sub-Gaussian norms, such that with probability at least $1-\eta$,
    \begin{equation}\label{eq:transformed_error_new}
    \|\tilde{\bm{\beta}}_{\scriptscriptstyle A}^{\,DP}
    -
    \tilde{\bm{\beta}}_{\scriptscriptstyle A}\|_2
    \le
    C_0\,
    \frac{
    d^{3/2}\Gamma_{A,\eta}(1+\delta_{A,\eta})^2
    }{
    \hat{\sigma}_{\scriptscriptstyle y,B}\,
    \sqrt{\rho_1}\,
    n_{\scriptscriptstyle A}\,
    L^{2}(1-\delta_{A,\eta})^{4}
    }\,
    \|\bm{\beta}^\star\|_2
    \|\Sigma\|_2
    \|\hat{\Sigma}_{\scriptscriptstyle B}^{-1/2}\|_2.
    \end{equation}
    Consequently, the mapped-back estimator error bound is
    \begin{equation}\label{eq:original_error_new}
		\|\bar{\bm{\beta}}_{\scriptscriptstyle A}^{\,DP}
		-
		\hat{\bm{\beta}}_{\scriptscriptstyle A}\|_2
		\le
		C_0
		\frac{d^{3/2}}{\sqrt{\rho_1}\,
		n_{\scriptscriptstyle A}\,}
		\frac{
		\Gamma_{A,\eta}(1+\delta_{A,\eta})^2
		}{
		L^2(1-\delta_{A,\eta})^4
		}
		\,
		\|\bm{\beta}^\star\|_2
		\|\Sigma\|_2
		\|\hat{\Sigma}_{\scriptscriptstyle B}^{-1}\|_2,
    \end{equation}
\end{enumerate}
\end{thm}

\begin{rem}[Decomposition of the utility bound]\label{rem:our_bound}
The bound in \eqref{eq:original_error_new} can be interpreted as the product of four factors:
\[
\underbrace{\frac{d^{3/2}}{\sqrt{\rho_1}\,n_{\scriptscriptstyle A}}}_{\text{generic DP rate}}
\cdot
\underbrace{\|\bm{\beta}^\star\|_2
\|\Sigma\|_2
\|\hat{\Sigma}_{\scriptscriptstyle B}^{-1}\|_2}_{\text{problem-dependent scale}}
\cdot
\underbrace{\frac{(1+\delta_{A,\eta})^2}{(1-\delta_{A,\eta})^4}}_{\text{concentration scale}}
\cdot
\underbrace{\frac{\Gamma_{A,\eta}}{L^2}}_{\textbf{preconditioned-geometry term}}.
\]
The first factor is the standard scaling induced by Gaussian perturbation in sufficient-statistics-based DP regression, reflecting the usual trade-off among dimension, privacy level, and private sample size. The second factor is problem-dependent and captures the signal strength together with the overall second-moment scale; under concentration, it behaves like $\|\bm{\beta}^\star\|_2\kappa(\Sigma)$, and is not the main component improved by our method. 

The third factor is a concentration term coming from the sample fluctuation of the second-moment estimation around its population counterpart. As $n_{\scriptscriptstyle A}$ increases, $\delta_{A,\eta}$ tends to zero, and this factor approaches $1$, so it represents only the finite-private-sample concentration cost.

The fourth factor is the method-specific term and captures the main geometric effect of public preconditioning. Here \(\Gamma_{A,\eta}\) comes from the clipping radius and its high-probability control: in the private-only geometry, this clip burden is tied to the anisotropic scale of the private covariates, whereas after public second-moment preconditioning it is reduced to a term depending essentially only on dimension, sample size, and failure probability. Meanwhile, \(L\) is the lower spectral bound of the transformed population second moment and serves as the key inverse-stability parameter: in the private-only geometry, the corresponding burden is governed by the possibly very small minimum eigenvalue of the private second moment, while after preconditioning \(L\) is driven close to \(1\) when \(n_{\scriptscriptstyle B}\) is sufficiently large. Therefore, the term \(\Gamma_{A,\eta}/L^2\) summarizes how public preconditioning simultaneously regularizes clip and stabilizes matrix inversion, replacing the explicit ill-conditioning burden of the original private covariance geometry. More detailed comparison is given in Subsection~\ref{subsec:comparison_private_only}.
\end{rem}

\begin{rem}[Remark on privacy budget allocation]
	The results of \textbf{Theorem~\ref{thm:DPPMTOLSE_main}} only show the privacy parameter $\rho_1 = \frac{\rho_{tot}}{2}$ of the second-moment matrix. Only the dependence on $\rho_1$ is displayed explicitly in the dominant term because, under equal or fixed budget splits, the contribution of the cross-moment perturbation is of lower order. A more refined analysis of optimal budget allocation between $\rho_1$ and $\rho_2$ is possible but omitted here.
\end{rem}

\subsection{Comparison with private-only sufficient statistics perturbation}\label{subsec:comparison_private_only}

To highlight the role of public preconditioning, we compare \textbf{Algorithm~\ref{alg:DPPMTOLSE}} with the standard private-only SSP estimator, \textbf{Algorithm \ref{alg:DPOLSE_baseline}} in Appendix \ref{app:baseline}. The baseline estimator is formally similar to SSP, but its clipping radius and its inverse stability are both governed by private-side scale and anisotropy. The corresponding error bound reflects this dependence.

\begin{thm}[Private-only SSP benchmark]\label{thm:DPOLSE_baseline}
With the same privacy setting as Algorithm~\ref{alg:DPPMTOLSE}, Algorithm~\ref{alg:DPOLSE_baseline} satisfies $\rho_{\mathrm{tot}}$-zCDP.

For $\eta\in(0,1)$, write
\[
\Lambda_{A,\eta}
:=
\left(\frac{\overline{\Tr}(\hat{\Sigma}_{\scriptscriptstyle A})
+
\log\!\frac{2n_{\scriptscriptstyle A}}{\eta}}{\lambda_{\min}(\hat{\Sigma}_{\scriptscriptstyle A})}\right) = \left(\bar{\kappa}(\hat{\Sigma}_{\scriptscriptstyle A}) + \|\hat{\Sigma}_{\scriptscriptstyle A}^{-1}\|_2\log\!\frac{2n_{\scriptscriptstyle A}}{\eta}\right),
\]
where $\overline{\Tr}(\hat{\Sigma}_{\scriptscriptstyle A})
=
\frac{1}{d}\sum_{j=1}^d \lambda_j(\hat{\Sigma}_{\scriptscriptstyle A})
$ denotes the average eigenvalue and 
$\bar{\kappa}(\hat{\Sigma}_{\scriptscriptstyle A})
=
\frac{1}{d}\sum_{j=1}^d
\frac{\lambda_j(\hat{\Sigma}_{\scriptscriptstyle A})}
{\lambda_{\min}(\hat{\Sigma}_{\scriptscriptstyle A})}
$ is an averaged condition-number.

Moreover, for every fixed $\eta\in(0,1)$, with probability at least $1-\eta$, if
\[
\frac{
d^{3/2}}{\sqrt{\rho_1}\,n_{\scriptscriptstyle A}}\Lambda_{A,\eta}
\le
\frac{1}{2},
\]
then there exists a constant $C_1>0$, depending only on the sub-Gaussian norms, such that
\begin{equation}\label{eq:baseline_error_new}
\|\hat{\bm{\beta}}_{\scriptscriptstyle A}^{\,DP}
-
\hat{\bm{\beta}}_{\scriptscriptstyle A}\|_2
\le
C_1\,
\frac{
d^{3/2}
}{
\sqrt{\rho_1}\,n_{\scriptscriptstyle A}
}\,\Lambda_{A,\eta}\,\frac{(1+\delta_{A,\eta})^2}{(1-\delta_{A,\eta})^2}\,
\|\bm{\beta}^\star\|_2\,
\|\Sigma\|\|\Sigma^{\scriptscriptstyle -1}\|,
\end{equation} where $\delta_{A,\eta}
:=C\!\left(\sqrt{\frac{d}{n_{\scriptscriptstyle A}}}+
\sqrt{\frac{\log(1/\eta)}{n_{\scriptscriptstyle A}}}\right)$.
\end{thm}

\begin{rem}[Proof of the private-only SSP]
	The proof follows the same steps as in Theorem \ref{thm:DPPMTOLSE_main}, specialized to the original geometry. In this situation, it's worth noting that the Gaussian perturbation matrix bound becomes $\|\mathbf{G}\|_2
=
O_{\mathbb P}\!\left(
\frac{d^{3/2}(\overline{\Tr}(\hat{\Sigma}_{\scriptscriptstyle \!A})+\log(2n_{\scriptscriptstyle \!A}/\eta))}
{\sqrt{\rho_1}\,n_{\scriptscriptstyle \!A}}
\right)$ as the naive clipping radius $\sqrt{\Tr(\hat{\Sigma}_{\scriptscriptstyle \!A})+d\log(2n_{\scriptscriptstyle \!A}/\eta)}$ and the smallest eigenvalue is the $\lambda_{\min} (\hat{\Sigma}_{\scriptscriptstyle \!A})$ rather than $L(1 - \delta_{A,\eta})^2$. The concentration scale $\frac{(1+\delta_{A,\eta})^2}{(1-\delta_{A,\eta})^2}$ is from the bounds of $\|\hat\Sigma_{\scriptscriptstyle A}\| \leq \|\Sigma\|(1+\delta_{A,\eta})^2$ and $\|\hat\Sigma_{\scriptscriptstyle A}^{\scriptscriptstyle -1}\| \leq \|\Sigma^{\scriptscriptstyle -1}\|(1-\delta_{A,\eta})^{-2}$.
\end{rem}

The comparison between \textbf{Theorem~\ref{thm:DPPMTOLSE_main}} and \textbf{Theorem~\ref{thm:DPOLSE_baseline}} is best understood by separating the utility bound into three components: a generic DP rate term, a problem-dependent scale term, and a geometry-dependent term. In both methods, the generic DP rate is of order
\[
\frac{d^{3/2}}{\sqrt{\rho_1}\,n_{\scriptscriptstyle A}},
\]
which reflects the usual trade-off among dimension, privacy level, and private sample size under Gaussian perturbation of sufficient statistics. In addition, both bounds contain problem-dependent scale factors involving the regression signal and second-moment magnitude, approximately as $\|\bm{\beta}^\star\| \kappa(\Sigma)$ . These terms are not the main source of improvement in our method. The third concentration terms, $\frac{(1+\delta_{A,\eta})^2}{(1-\delta_{A,\eta})^4}$ and $\frac{(1+\delta_{A,\eta})^2}{(1-\delta_{A,\eta})^2}$, tend to $1$ as the private-sample $n_{\scriptscriptstyle A}$ increases with a similar rate.

The essential difference lies in the geometry-dependent part. For the private-only SSP estimator, the inverse-stability requirement is governed by
\[
\Lambda_{A,\eta}
=
\left(
\bar{\kappa}(\hat{\Sigma}_{\scriptscriptstyle A})
+
\|\hat{\Sigma}_{\scriptscriptstyle A}^{-1}\|_2
\log\!\frac{2n_{\scriptscriptstyle A}}{\eta}
\right),
\]
which depends explicitly on the anisotropy of the private second-moment matrix through both the averaged condition measure $\bar{\kappa}(\hat{\Sigma}_{\scriptscriptstyle A})$ and the small-eigenvalue factor $\|\hat{\Sigma}_{\scriptscriptstyle A}^{-1}\|_2$. By contrast, in the transformed estimator this burden is replaced by
\[
\frac{\Gamma_{A,\eta}}{L^2} = \frac{1+\log\!\frac{2n_{\scriptscriptstyle A}}{\eta}}{\Big(1 + C\Big(\sqrt{\frac{d}{n_{\scriptscriptstyle \!B}}} + \sqrt{\frac{\log(1/\eta)}{n_{\scriptscriptstyle \!B}}}\Big)\Big)^{-4}}
\]
Here $\Gamma_{A,\eta}$ and $L$ summarize the geometry after public preconditioning: $\Gamma_{A,\eta}$ contains the upper spectral bound and the clipping radius and $L$ gives a lower spectral bound for the transformed covariance. Different to $\Lambda_{A,\eta}$, they are independent of $\Sigma$.

To make the comparison more transparent, consider the idealized large-sample regime in which $\hat{\Sigma}_{\scriptscriptstyle A}$ concentrates around $\Sigma$. Under this heuristic simplification, the private-only stability condition behaves like
\[
\frac{d^{3/2}}{\sqrt{\rho_1}\,n_{\scriptscriptstyle A}}
\left(
\bar{\kappa}(\Sigma)
+
\|\Sigma^{-1}\|_2\log\!\frac{2n_{\scriptscriptstyle A}}{\eta}
\right),
\]
whereas the transformed stability condition behaves like
\[
\frac{d^{3/2}}{\sqrt{\rho_1}\,n_{\scriptscriptstyle A}}
\frac{
1+\log\!\frac{2n_{\scriptscriptstyle A}}{\eta}
}{
L(1-\delta_{A,\eta})^2
}.
\]
When the public sample is sufficiently large, $L$ is close to $1$; when the private sample is sufficiently large, $\delta_{A,\eta}$ is nearly zero. This distinction is especially important when the original regression problem is poorly conditioned. For instance, if two problems have the same dimension and privacy level, but one satisfies $\lambda_{\min}(\Sigma)\asymp 1$ and $\bar{\kappa}(\Sigma)\asymp 1$, whereas the other satisfies $\lambda_{\min}(\Sigma)\asymp 10^{-2}$ and $\bar{\kappa}(\Sigma)\asymp 10^2$, then the private-only stability requirement may be inflated by roughly two orders of magnitude. But our method largely avoids this explicit ill-conditioning burden through public preconditioning.

\section{Experiments}\label{sec:experiment}

We evaluate the proposed method from three complementary perspectives: 
(i) how the estimation error changes with the private sample size; 
(ii) how the gain depends on the conditioning of the regression geometry; and 
(iii) how the quality of the public second-moment estimator affects performance. 
Throughout, we compare our method \textbf{DP-PMTOLSE} with the most directly comparable baseline in Appendix~\ref{app:baseline}, simply referring to \textbf{Algorithm~\ref{alg:DPOLSE_baseline}} as \textbf{DP-OLSE}. This comparison isolates the effect of \emph{public second-moment preconditioning} within the same SSP framework. We also compare the DP gradient work, \textbf{Algorithm 2} in \citet{nasr2023effectively}, in Appendix~\ref{app:dope_compare}. 

Unless otherwise stated, all reported results are averaged over repeated Monte Carlo trials and reported as the box plots. In addition, we report key geometry diagnostics, including the minimum eigenvalue and averaged condition-number proxy of the empirical second-moment matrices before and after transformation. Our primary error metric is the Euclidean parameter error. In synthetic experiments, we report the error between DP estimator and the real parameter
\[
\|\hat{\bm \beta}^{DP}-\bm \beta^\star\|_2,
\]
while in real-data experiments we report the error between DP estimator and the estimator from the full data 
\[
\|\hat{\bm \beta}^{DP}-\hat{\bm \beta}_{\mathrm{full}}\|_2,
\]
where \(\hat{\bm \beta}_{\mathrm{full}}\) is the non-private OLS estimator computed from the full available dataset.

\subsection{Synthetic experiments}\label{subsec:synthetic_experiment}

\textbf{Synthetic model and implementation details.} We consider Gaussian linear regression with dimension \(d=10\). For each trial, the private and public covariates are sampled independently from a common Gaussian distribution with the covariance matrix $\Psi$, where the second-moment matrix is equal to the covariance matrix $\Psi$ when the mean $\bm \mu = \bm 0$,
\[
\mathbf{x} \sim \mathcal{N}(\bm 0,\Psi),
\]
and the response is generated from
\[
y = \mathbf{x}^\top \bm\beta^\star + \bm \omega,
\qquad
\bm \omega \sim \mathcal{N}(0,0.05^2).
\]
To isolate the effect of geometry, we fix the population parameter \(\bm \beta^\star\) across trials and either fix or vary the covariance structure depending on the experiment. The target covariance geometry is controlled through the condition number of \(\Psi\), and all experiments use failure probability \(\eta=0.05\). 

In the preconditioned method, the public second-moment matrix is computed from the public sample and used to form the transformed design. The transformed covariates and responses are then truncated according to the dimension-scale radii derived in Section~\ref{sec:Main Theoretical Results}. In the private-only baseline, truncation is performed directly in the original coordinates using the private empirical scale. Both methods split the total privacy budget equally between the perturbed second-moment matrix and the perturbed cross-moment vector.

\subsubsection{Effect of private sample size under fixed geometry}

We first fix the population geometry and vary the private sample size. Specifically, we set \(d=10\), \(n_B=50\), \(\kappa(\Psi)=10\), \(\lambda_{\max} (\Psi) = 1\) and fix \(\bm \beta^\star\) with $ \|\bm \beta^\star\|_2= 4.1879$. We then vary
\[
n_A \in \{3\times 10^4,\; 5\times 10^4,\; 7\times 10^4,\; 10^5,\; 5\times 10^5\},
\]
and consider total privacy budgets \(\rho \in \{1,10,50\}\). Each configuration is repeated over \(300\) independent trials.

\begin{figure}[H]
    \centering
    \includegraphics[width=\textwidth]{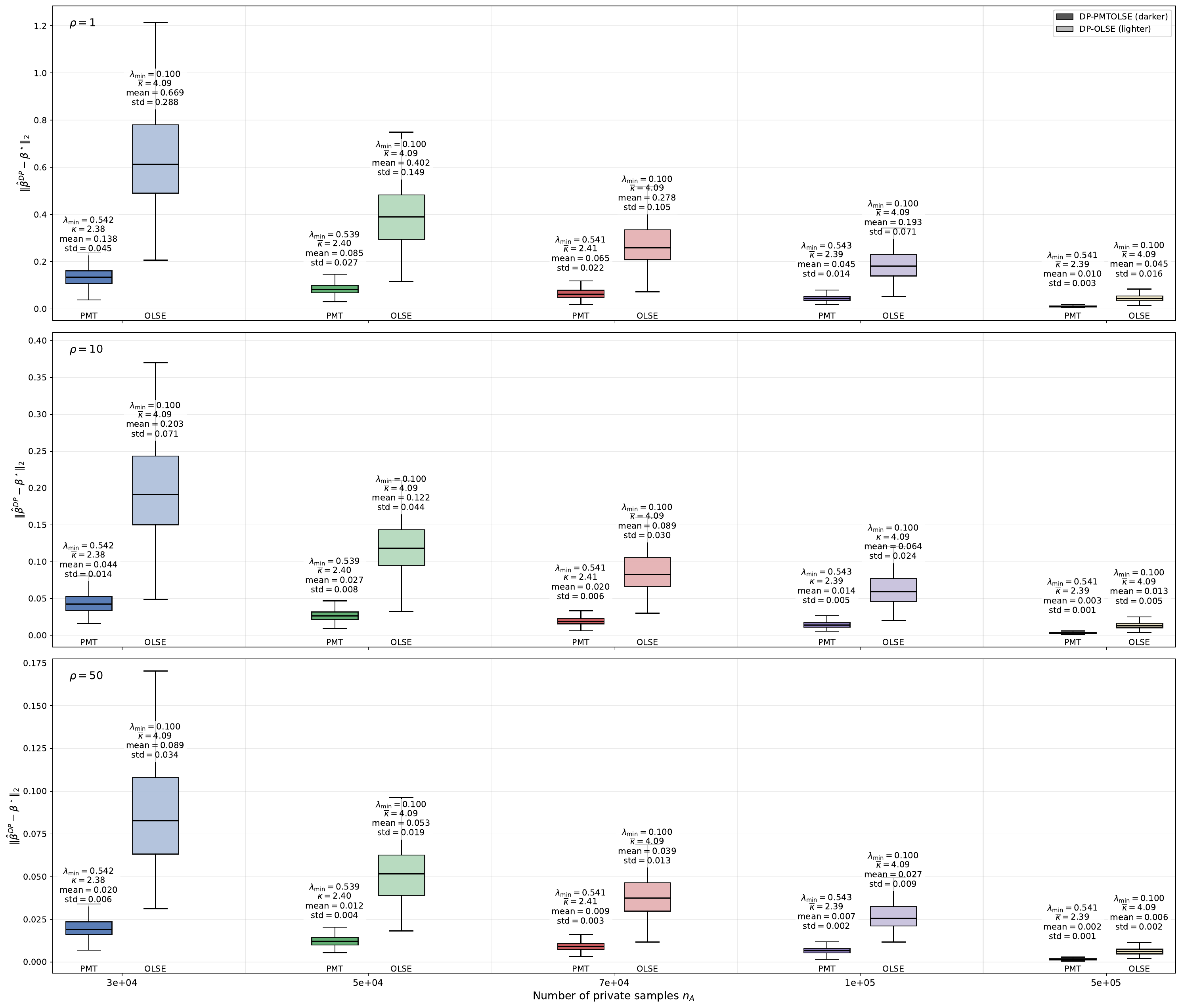}
    \caption{Synthetic experiment under fixed geometry: mean \(\pm\) standard deviation of \(\|\hat{\bm \beta}^{DP}-\bm \beta^\star\|_2\) over 300 trials, for varying private sample size \(n_A\) and privacy budget \(\rho\).}
    \label{fig:synthetic_private_n}
\end{figure}

Figure~\ref{fig:synthetic_private_n} shows the mean \(\pm\) standard deviation of \(\|\hat{\bm \beta}^{DP}-\bm \beta^\star\|_2\) for both methods. Several patterns are clear. First, for both methods the error decreases as \(n_A\) increases, which is consistent with the theoretical \(1/n_A\)-type scaling in the utility bounds. Second, for each privacy level, \textbf{DP-PMTOLSE} consistently outperforms \textbf{DP-OLSE}; the gap is still pronounced in the low-privacy regime (\(\rho=50\)), even through the private-only estimator suffers less from noise. Third, the variability of \textbf{DP-PMTOLSE} is substantially smaller, indicating improved numerical robustness in addition to improved mean accuracy.

For example, under \(\rho=1\), the \textit{mean} error decreases from approximately \(0.138\) to \(0.010\) for \textbf{DP-PMTOLSE} as \(n_A\) grows from \(3\times 10^4\) to \(5\times 10^5\), while the corresponding \textbf{DP-OLSE} error decreases from about \(0.669\) to \(0.045\). \textbf{DP-PMTOLSE} always holds the low-level error standard deviations (\textit{std}), but the stability of \textbf{DP-OLSE} depends on \(n_A\) seriously. About the empirical covariance matrix, the public preconditioning raises the minimum eigenvalue $\lambda_{\min}$ and decreases the averaged condition-number $\bar{\kappa}$. Similar relative gains persist for \(\rho=10\) and \(\rho=50\). These results are consistent with the theory that public preconditioning improves the stability and accuracy of the SSP linear regression.

\subsubsection{Effect of second-moment martix conditioning}

We next examine how the benefit of public preconditioning changes with the condition number of the regression geometry. Here we fix \(d=10\), \(n_A=10^4\), \(n_B=50\), \(\lambda_{\max} (\Psi) = 1\), and \(\bm \beta^\star\) with $\|\bm \beta^\star\|_2= 4.1879$, and vary
\[
\kappa(\Psi)\in\{1,5,10,20,30\}.
\]
We consider \(\rho\in\{1,50\}\) and repeat each setting \(300\) times.

\begin{figure}[H]
    \centering
    \includegraphics[width=\textwidth]{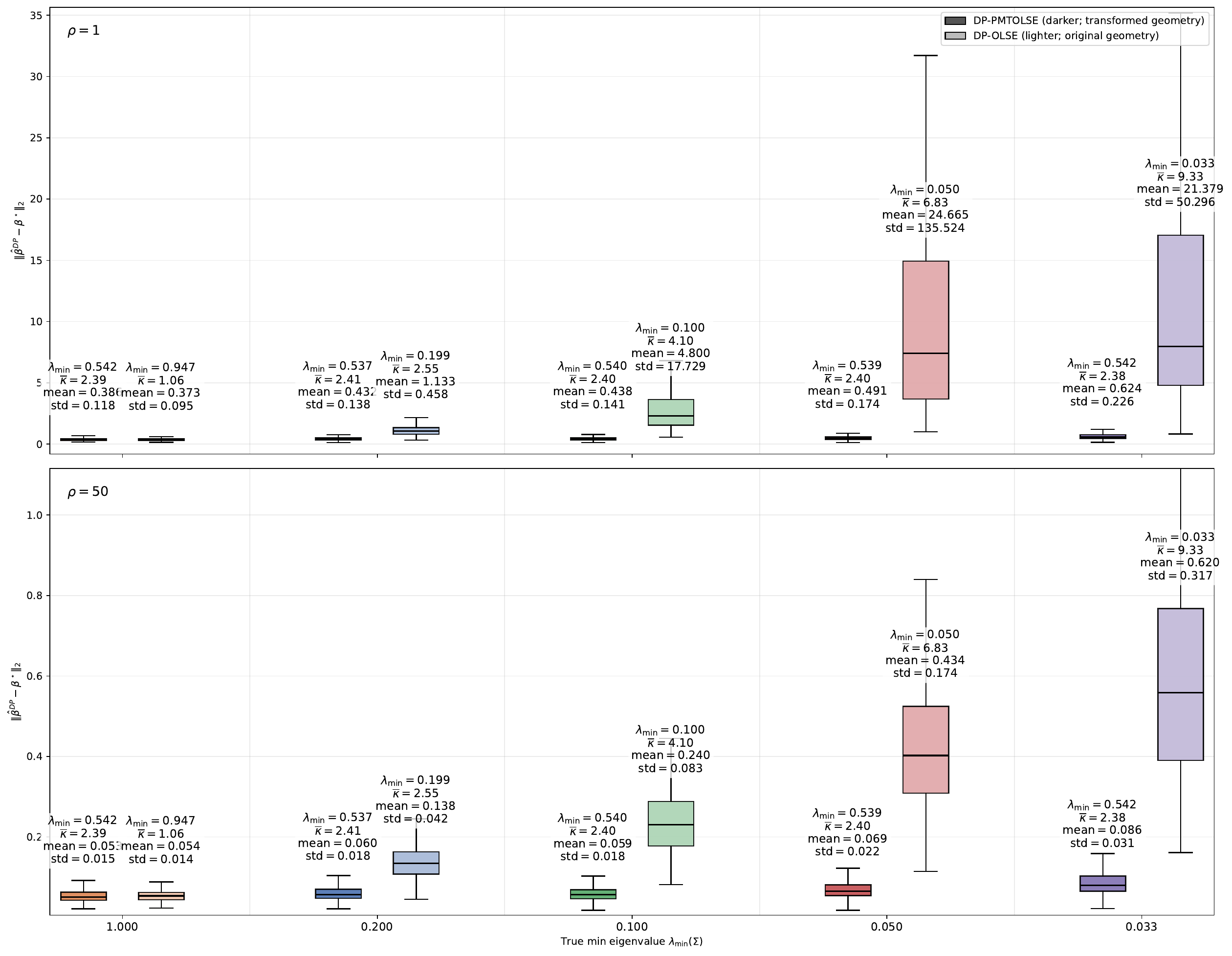}
    \caption{Synthetic experiment with varying condition number: distribution of \(\|\hat{\beta}^{DP}-\beta^\star\|_2\) over 300 trials, for different target condition numbers \(\kappa(\Psi)\) and privacy budgets \(\rho\).}
    \label{fig:synthetic_kappa}
\end{figure}

Figure~\ref{fig:synthetic_kappa} reports the box plot of \(\|\hat{\bm \beta}^{DP}-\bm \beta^\star\|_2\) across trials. When the geometry is the isotropic (\(\kappa(\Psi)=1\)), the two methods perform similarly, especially under the weaker privacy regime \(\rho=50\). This is expected: when the original geometry is already well conditioned, there is little room for improvement through preconditioning. In contrast, as \(\kappa(\Psi)\) increases, the performance of \textbf{DP-OLSE} deteriorates rapidly, with both its mean error and variance becoming very large, while \textbf{DP-PMTOLSE} remains stable.

This trend is particularly sharp under \(\rho=1\). For instance, when \(\kappa(\Psi)=20\) or \(\lambda_{\min} = 0.05\), the mean error of \textbf{DP-OLSE} exceeds \(20\) with extremely large variability, whereas the mean error of \textbf{DP-PMTOLSE} remains below \(0.5\). The corresponding geometry diagnostics show that the transformed second-moment matrix is much closer to isotropic than the original private second moment, which directly supports the theoretical message of Section~\ref{sec:Main Theoretical Results}: the main advantage of our method appears in anisotropic and ill-conditioned regimes.

\subsubsection{Effect of the public sample size}

Finally, we study how the size of the public sample affects the quality of the preconditioner. We fix \(d=10\), \(n_A=10^4\), \(\lambda_{\max} (\Psi) = 1\), \(\bm \beta^\star\) with $\|\bm \beta^\star\|_2= 4.1879$, and vary
\[
n_B \in \{11,20,100,1000\},
\qquad
\kappa(\Psi)\in\{1,10,50\},
\qquad
\rho\in\{1,10\}.
\]
Each setting is repeated \(500\) times.

\begin{figure}[H]
    \centering
    \includegraphics[width=\textwidth]{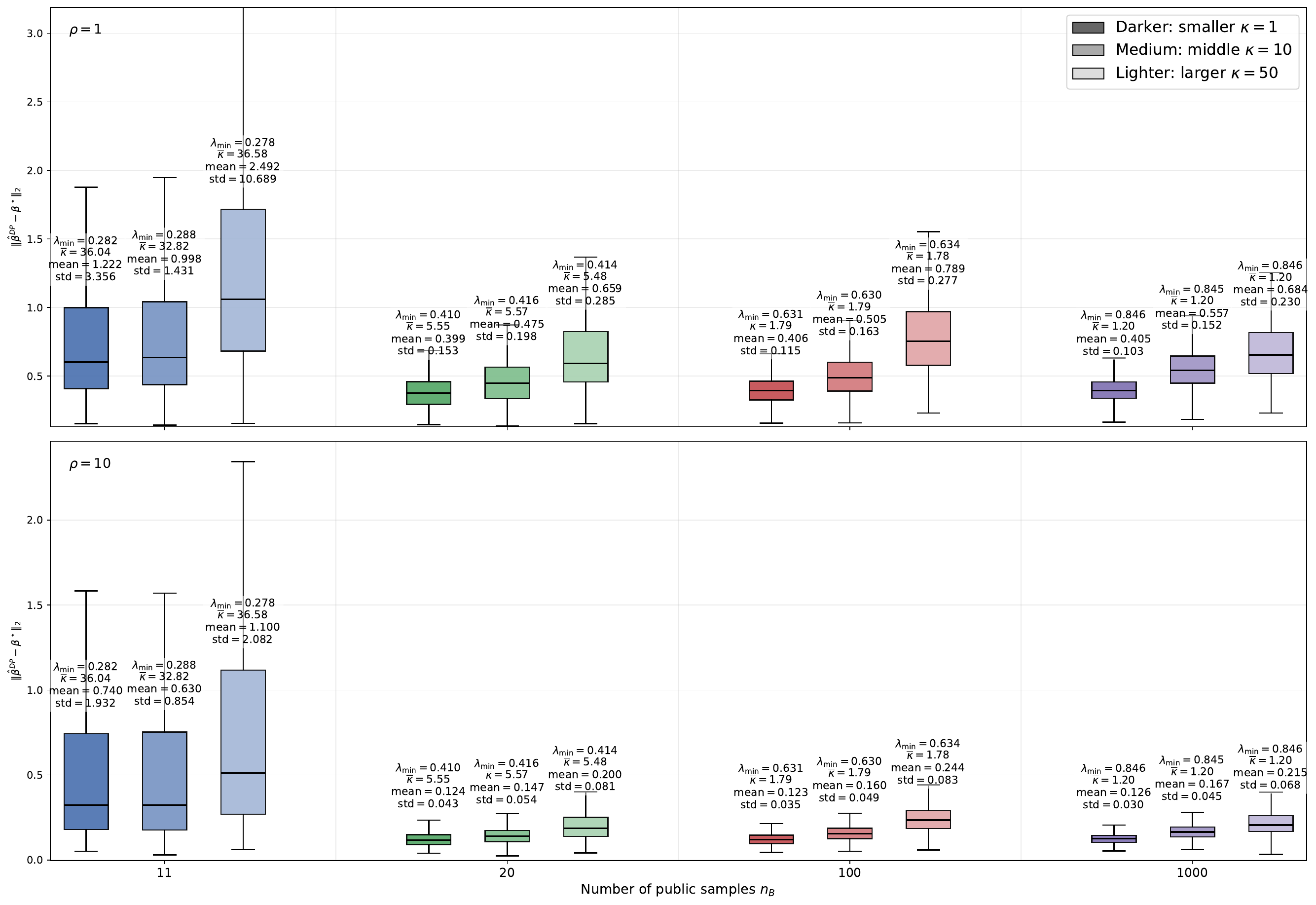}
    \caption{Synthetic experiment for public sample size: distribution of \(\|\hat{\bm \beta}^{DP}-\bm \beta^\star\|_2\) for \textbf{DP-PMTOLSE} over 500 trials, under varying \(n_B\), target condition number \(\kappa(\Psi)\), and privacy budget \(\rho\).}
    \label{fig:synthetic_public_n}
\end{figure}

Figure~\ref{fig:synthetic_public_n} shows the error of \textbf{DP-PMTOLSE} as a function of \(n_B\). The results reveal a threshold phenomenon: once the public sample size is slightly larger than the dimension, the performance becomes much more stable, which is consistent with the requirement that \(\hat{\Sigma}_B\) be invertible and spectrally well behaved. Moving from \(n_B=11\) to \(n_B=20\) produces a substantial reduction in both error and variability; further increases in \(n_B\) continue to improve performance, but with diminishing returns. This is again in line with the theory: a larger public sample improves the approximation of the population second-moment structure, thereby driving the transformed geometry closer to isotropy and increasing the lower spectral bound \(L\).

The dependence on the condition number \(\kappa(\Psi)\) is also shown in the experiments. When the target condition number $\kappa$ increases from $1$ to $50$, the estimation errors also become larger. These results corresponds to our theoretical bound, seeing the \textit{problem-dependent scale} approximated as $\|\bm{\beta^\star}\|\kappa(\Sigma)$ in \textbf{Remark \ref{rem:our_bound}}.

\subsection{Real-data experiments}\label{subsec:real_experiment}

We complement the synthetic experiments with two real-data studies. The first uses a continuous-response dataset that is directly aligned with linear regression; the second uses the white-wine quality dataset to test the method in a more ill-conditioned practical setting. All datasets are standardized as the mean $0$ and the variance $1$, in order to eliminate the influence of the location and the scale.

\subsubsection{Combined Cycle Power Plant}

Our first real-data benchmark is the UCI Combined Cycle Power Plant dataset, which contains \(9568\) observations and \(d=4\) continuous covariates. The response is continuous, making this dataset particularly appropriate for ordinary least squares regression. We compute the full-data OLS estimator \(\hat{\bm \beta}_{\mathrm{full}}\), which serves as the non-private target.

The full-data empirical second moment has minimum eigenvalue \(0.1026\) and averaged condition-number \(\bar{\kappa} = 9.75\), indicating moderate but nontrivial anisotropy. In each trial, we randomly select \(50\) public samples and use the remainder as the private dataset. From the private dataset, we draw private subsets of sizes corresponding to fractions $\{30\%,\ 50\%,\ 70\%,\ 100\%\}$ of the available private data, and evaluate both \textbf{DP-OLSE} and \textbf{DP-PMTOLSE} under \(\rho\in\{1,10\}\). Each setting is repeated \(300\) times.

\begin{figure}[H]
    \centering
    \includegraphics[width=\textwidth]{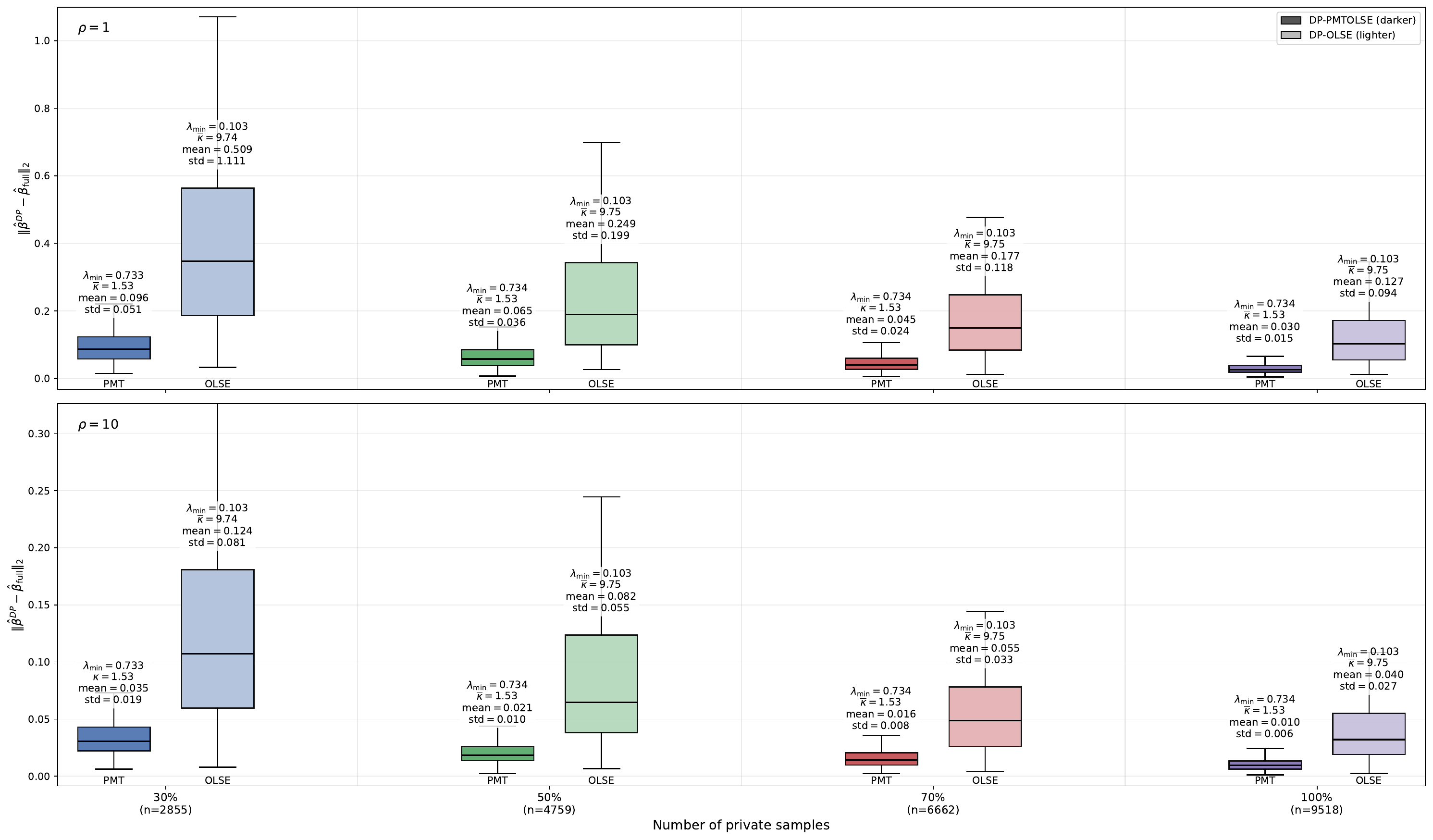}
    \caption{Combined Cycle Power Plant dataset: mean \(\pm\) standard deviation of \(\|\hat{\bm \beta}^{DP}-\hat{\bm \beta}_{\mathrm{full}}\|_2\) over 300 trials, for varying private sample size fractions and privacy budget \(\rho\).}
    \label{fig:ccpp_real}
\end{figure}

Figure~\ref{fig:ccpp_real} shows that \textbf{DP-PMTOLSE} consistently yields lower error than the private-only baseline. Under \(\rho=1\), for example, the mean error drops from \(0.509\) to \(0.096\) at the \(30\%\) private fraction and from \(0.127\) to \(0.030\) at the full private size. The improvement remains substantial under \(\rho=10\). These results show that even in a moderately conditioned dataset, public second-moment preconditioning provides a clear utility gain.

\subsubsection{White-wine quality data}

We further evaluate the method on the UCI Wine Quality dataset with $11$ features, using only the white-wine subset. Its empirical second moment is substantially more ill conditioned than in the previous real-data example. Following a mild preprocessing rule, we remove only $9$ extreme upper-tail outliers in the covariates, retain \(4889\) samples, and standardize both covariates and response. The resulting empirical second moment has minimum eigenvalue \(0.0145\) and averaged condition-number \(69.03\), making this dataset a useful stress test for numerical stability.

In each trial, we again fix the public sample size to \(50\), and use private subsets corresponding to fractions $\{40\%,\ 60\%,\ 80\%,\ 100\%\}$
of the remaining private datasets. We consider \(\rho\in\{1,50\}\) and repeat each setting \(300\) times.

\begin{figure}[H]
    \centering
    \includegraphics[width=\textwidth]{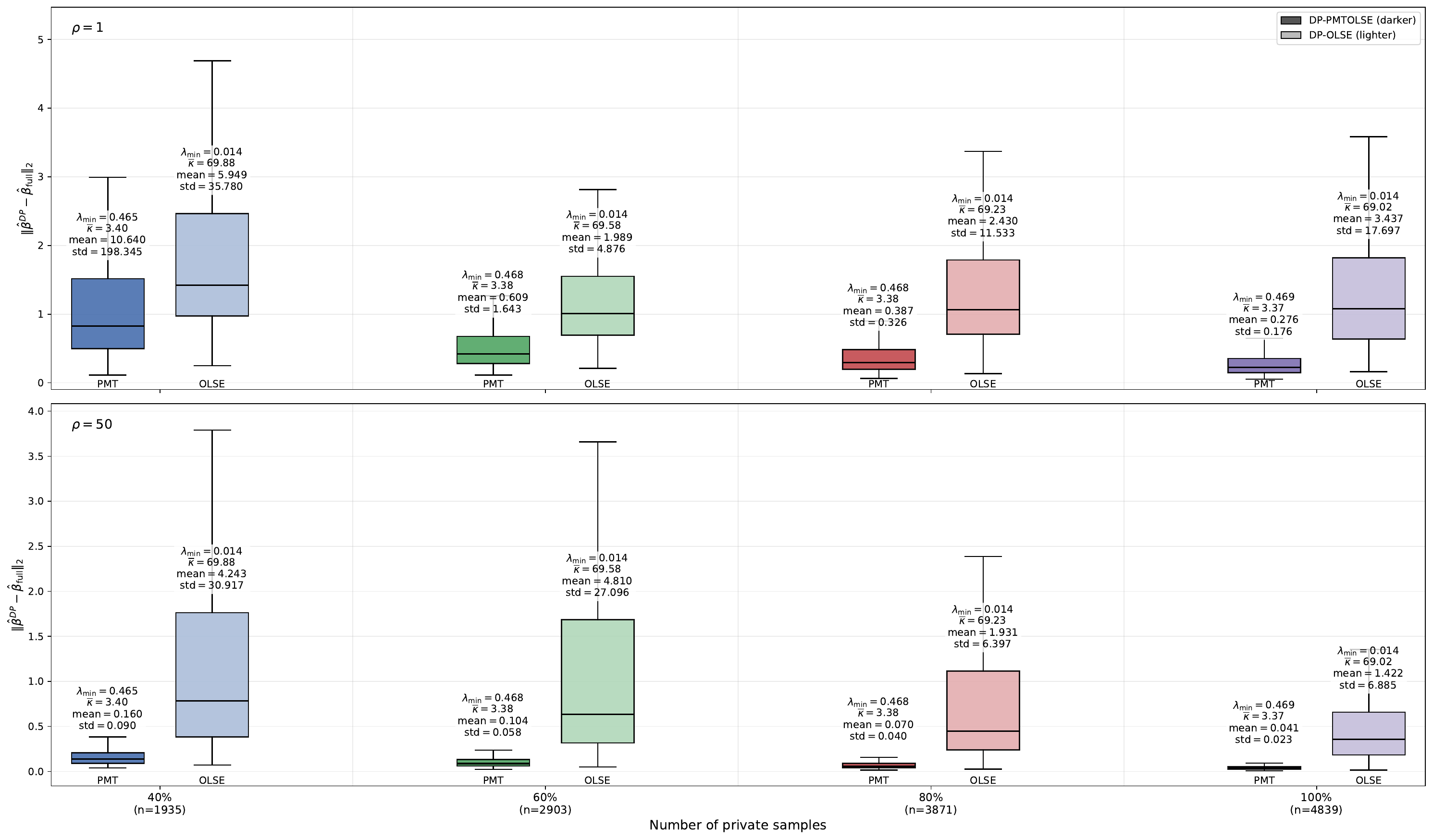}
    \caption{White-wine quality dataset: mean \(\pm\) standard deviation of \(\|\hat{\bm \beta}^{DP}-\hat{\bm \beta}_{\mathrm{full}}\|_2\) over 300 trials, for varying private sample size fractions and privacy budget \(\rho\).}
    \label{fig:wine_real}
\end{figure}

Figure~\ref{fig:wine_real} shows that the advantage of \textbf{DP-PMTOLSE} is even more pronounced in this ill-conditioned setting. Under \(\rho=1\), the private-only estimator displays very large variability and, in some settings, extremely unstable behavior, whereas the preconditioned estimator remains substantially more accurate and stable. For example, at the full private size, the mean error decreases from \(3.43\) for \textbf{DP-OLSE} to \(0.276\) for \textbf{DP-PMTOLSE}. Under \(\rho=50\), the corresponding errors decrease from \(1.422\) to \(0.041\). This strong empirical gap is precisely the regime highlighted by our theory: when the original private second moment is poorly conditioned, public preconditioning significantly improve the estimation accuracy and reduces the instability caused by perturbing the sufficient statistics.

\subsection{Discussion of the empirical findings}

The experiments support three conclusions. First, \textbf{DP-PMTOLSE} consistently improves over the private-only SSP baseline across both synthetic and real-data settings. Second, the gain is modest when the design is already close to isotropic, but becomes substantial when the covariance geometry is ill conditioned. Third, the quality of the public preconditioner matters: once the public sample size is large enough to estimate the second-moment geometry reliably, the transformed-domain estimator becomes both more accurate and more stable. 

These findings match the theory developed in Section~\ref{sec:Main Theoretical Results}. The proposed method does not change the generic privacy-induced rate, but it substantially improves the geometry-dependent part of the utility bound by reducing truncation burden and stabilizing the inverse of the privatized second-moment matrix.

\section{Conclusion and Future Work}\label{sec:conclusion}
In this work, we propose a method that transforms private data for differentially private linear regression by utilizing the second-moment of public data. We demonstrate the difference between our method and the conventional SSP regression and theoretically present an error bound between our method and the non-DP. These theoretical findings provide an explanation for the improved accuracy and robustness of our approach. We validate the methods on both synthetic and real datasets. The experiments show that the second-moment matrix estimation using tiny amounts of public data can greatly enhance DP-OLSE. 

To isolate the role of public-moment preconditioning itself, we develop the analysis in a benchmark regime where public and private covariates share the same second-order structure. This benchmark allows us to cleanly characterize how public geometry improves conditioning, sensitivity control, and utility. In practice, public and private samples may differ because of sampling bias, temporal drift, or population shift. Extending the framework to heterogeneous public/private sources is an important but substantially more difficult problem, since the effectiveness of the transformation then depends on how well the public second-moment structure approximates its private counterpart. We leave this problem for future work.
\section{Acknowledgement}
This work was supported by the National Natural Science Foundation of China (Grant Number: 12326615).

\section*{Appendix}

\begin{appendices}

	\section{Technical tools}\label{app:technical_tools}

	This appendix collects the auxiliary results used in the main proofs. They are standard concentration and perturbation tools for sub-Gaussian vectors, sample second-moments, inverse matrix perturbation, and Gaussian noise. We place them here to keep the main text focused on the method and its interpretation.

	\subsection{Norm concentration and sample second-moments}

	The first lemma gives a high-probability norm bound for a non-isotropic sub-Gaussian vector. It is used to motivate clip radius.

	\begin{lem}[Norm concentration for non-isotropic sub-Gaussian vectors]\label{lem:nonisotropic_norm}
	Let $\mathbf{x}=(x_1,\ldots,x_d)^\top\in\mathbb{R}^d$ be a mean-zero sub-Gaussian random vector with second-moment matrix
	\[
	\Sigma=\mathbb{E}(\mathbf{x}\mathbf{x}^\top).
	\]
	Assume $\max_{1\le j\le d}\|x_j\|_{\psi_2}\le K$. Then $\|\mathbf{x}\|_2^2$ is sub-exponential, and there exists an absolute constant $c>0$ such that for every $t>0$,
	\[
	\mathbb{P}\!\left(
	\bigl|\|\mathbf{x}\|_2^2-\Tr(\Sigma)\bigr|
	\ge t
	\right)
	\le
	2\exp\!\left(-\frac{ct}{dK^2}\right).
	\]
	Consequently, for every $\eta\in(0,1)$, with probability at least $1-\eta$,
	\[
	\|\mathbf{x}\|_2^2
	\le
	\Tr(\Sigma)
	+
	\frac{dK^2}{c}\log\!\left(\frac{2}{\eta}\right).
	\]
	\end{lem}

	The next lemma gives singular-value control for a random matrix with independent sub-Gaussian rows.

	\begin{lem}[Singular-value bounds for sub-Gaussian matrices {\citep{vershynin2010introduction}}]\label{lem:singular_Bound}
	Let $\mathbf{X}\in\mathbb{R}^{n\times d}$ have independent rows $\mathbf{x}_1,\ldots,\mathbf{x}_n$, each distributed as a sub-Gaussian random vector in $\mathbb{R}^d$ with second-moment matrix $\Sigma$. Then there exist constants $C,c>0$, depending only on the sub-Gaussian norm, such that for every $\eta\in(0,1)$, with probability at least $1-2\eta$,
	\[
	\sqrt{\lambda_{\min}(\Sigma)}
	\left(
	\sqrt{n}-C\sqrt{d}-\sqrt{\frac{1}{c}\log\frac{1}{\eta}}
	\right)
	\le
	s_{\min}(\mathbf{X}),
	\]
	and
	\[
	s_{\max}(\mathbf{X})
	\le
	\sqrt{\lambda_{\max}(\Sigma)}
	\left(
	\sqrt{n}+C\sqrt{d}+\sqrt{\frac{1}{c}\log\frac{1}{\eta}}
	\right),
	\]
	where $s_{\min}(\mathbf{X})$ and $s_{\max}(\mathbf{X})$ denote the smallest and largest singular values of $\mathbf{X}$.
	\end{lem}

	As a direct consequence, sample second-moments concentrate around the identity in the isotropic case.

	\begin{cor}\label{cor:secondmoment}
	Let $\mathbf{X}\in\mathbb{R}^{n\times d}$ have independent isotropic sub-Gaussian rows, i.e.\ $\mathbb{E}(\mathbf{x}_i\mathbf{x}_i^\top)=\mathbf{I}_d$. Then for every $\eta\in(0,1)$, with probability at least $1-2\eta$,
	\[
	\lambda_{\max}\!\left(\frac{1}{n}\mathbf{X}^\top\mathbf{X}\right)
	\le
	\left(
	1+C\!\left(\sqrt{\frac{d}{n}}+\sqrt{\frac{\log(1/\eta)}{n}}\right)
	\right)^2,
	\]
	and
	\[
	\lambda_{\min}\!\left(\frac{1}{n}\mathbf{X}^\top\mathbf{X}\right)
	\ge
	\left(
	1-C\!\left(\sqrt{\frac{d}{n}}+\sqrt{\frac{\log(1/\eta)}{n}}\right)
	\right)^2,
	\]where $C$ is an absolute constant. 
	\end{cor}

	\subsection{Inverse perturbation bound}

	The following perturbation lemma is the key tool for controlling the inverse of a privatized second-moment matrix.

	\begin{lem}[Inverse perturbation bound]\label{lem:inversebound}
	Let $\Sigma$ be an invertible square matrix and let $\mathbf{G}$ be a perturbation matrix. Write
	\[
	\kappa(\Sigma)=\|\Sigma\|\,\|\Sigma^{-1}\|
	\]
	for the condition number induced by the operator norm. Then
	\[
	\frac{\|(\Sigma+\mathbf{G})^{-1}-\Sigma^{-1}\|}{\|(\Sigma+\mathbf{G})^{-1}\|}
	\le
	\kappa(\Sigma)\frac{\|\mathbf{G}\|}{\|\Sigma\|}.
	\]
	If
	\[
	\|\Sigma^{-1}\|\,\|\mathbf{G}\|<1,
	\]
	then $\Sigma+\mathbf{G}$ is invertible and
	\[
	\|(\Sigma+\mathbf{G})^{-1}\|
	\le
	\frac{\|\Sigma^{-1}\|}{1-\|\Sigma^{-1}\|\,\|\mathbf{G}\|},
	\]
	as well as
	\[
	\frac{\|(\Sigma+\mathbf{G})^{-1}-\Sigma^{-1}\|}{\|\Sigma^{-1}\|}
	\le
	\frac{\kappa(\Sigma)\frac{\|\mathbf{G}\|}{\|\Sigma\|}}
	{1-\kappa(\Sigma)\frac{\|\mathbf{G}\|}{\|\Sigma\|}}.
	\]
	\end{lem}

	\subsection{Gaussian matrix and vector concentration}\label{app:Gaussian_tail}

	We also need concentration bounds for the Gaussian perturbations added by the privacy mechanism.

	\begin{lem}[Symmetric Gaussian matrix norm bound {\citep{vershynin2020high}}]\label{lem:GUEtail}
	Let $\mathbf{W}\in\mathbb{R}^{d\times d}$ be a symmetric Gaussian matrix whose upper-triangular entries are independent and distributed as $\mathcal{N}(0,\sigma^2)$. Then there exist absolute constants $C,c>0$ such that for every $t\ge 0$,
	\[
	\mathbb{P}\!\left(\|\mathbf{W}\|_2 \ge \sigma (2\sqrt{d} + t)\right)
	\le
	2\exp(-ct^2).
	\]
	In particular, for every $\eta\in(0,1)$, there exist an absolute constant $C' >0$, with probability at least $1-\eta$,
	\[
	\|\mathbf{W}\|_2
	\le
	C'\sigma\!\left(\sqrt{d} +\sqrt{\log(1/\eta)}\right).
	\] 
	\end{lem}

	\begin{lem}[Gaussian vector norm bound {\citep{laurent2000adaptive}}]\label{lem:gaussiannorm}
	Let $\mathbf{z}\sim \mathcal{N}(0,\sigma^2\mathbf{I}_d)$. Then for every $\eta\in(0,1)$,
	\[
	\mathbb{P}\!\left(
	\|\mathbf{z}\|_2^2
	\ge
	\sigma^2\Bigl(d+2\sqrt{d\log(1/\eta)}+2\log(1/\eta)\Bigr)
	\right)
	\le
	\eta.
	\]
	\end{lem}

	\section{Proofs for Lemmas}
		\subsection{Proof of Lemma \ref{lem:nonisotropic_norm}}
				
			\begin{prf}
			For simplicity, we assume that $K \geq 1$. Since the each coordinate $x_i$ is sub-gaussian, $x_i^2$ is sub-exponential, and more precisely 
			\begin{equation*}
			\begin{aligned}
				\big\|\|\mathbf{x}\|_2^2\big\|_{\psi_1} &= \| \sum_{i=1}^{d} x_i^2\|_{\psi_1}\\
				&\leq \sum_{i=1}^{d} \|x_i^2\|_{\psi_1} \\
				&\leq d\max_i \|x^2_i\|_{\psi_1} \\
				&= d \max_i \|x_i\|_{\psi_2}^2.
			\end{aligned}
			\end{equation*}Hence, $\|\mathbf{x}\|_2^2-\mathbb E\|\mathbf{x}\|_2^2$ is sub-exponential with $\psi_1$-norm bounded by a constant multiple of $dK^2$.
			Then, we compute the expectation of $\|\mathbf{x}\|_2^2$
			\begin{equation*}
			\begin{aligned}
				\mathbb{E} \|\mathbf{x}\|_2^2 &= \mathbb{E} \mathbf{x}^{\scriptscriptstyle \!\!\top} \mathbf{x}\\
				& =\mathbb{E}  \Tr(\mathbf{x}^{\scriptscriptstyle \!\!\top} \mathbf{x}) \\
				&= \mathbb{E} \Tr(\mathbf{x} \mathbf{x}^{\scriptscriptstyle \!\!\top}) \\
				&= \Tr(\mathbb{E} \mathbf{x} \mathbf{x}^{\scriptscriptstyle \!\!\top}) \\
				&= \Tr(\Sigma) \\
				&= \sum_{i=1}^{d} \lambda_i.
			\end{aligned}
			\end{equation*}
			Reconsider the tail bound of the centered sub-exponential random, we have 
			\begin{equation*}
				\mathbb{P}\Big[\Big|\|\mathbf{x}\|_2^2 - \mathbb{E}\|\mathbf{x}\|_2^2\Big| > t \Big] \leq 2\exp\big(-\frac{ct}{dK^2}\big),
			\end{equation*}where $K = \max_i \|x_i\|_{\psi_2}$ and $c$ is an absolute constant. 	$\hfill\square$
			\end{prf}

		\subsection{Proof of Lemma \ref{lem:precondition_spectrum}}
			\begin{prf}
				The goal 
					\[
					L\,\mathbf{I}_d
					\preceq
					\tilde{\Sigma}
					\preceq
					U\,\mathbf{I}_d
					\]
				is equal to 
				\begin{equation}\label{eq:LU_equal}
					L\Sigma^{\scriptscriptstyle \!-\!1\!/\!2}\hat{\Sigma}_{\scriptscriptstyle \!B}\Sigma^{\scriptscriptstyle \!-\!1\!/\!2} \preceq \mathbf{I} \preceq U\Sigma^{\scriptscriptstyle \!-\!1\!/\!2}\hat{\Sigma}_{\scriptscriptstyle \!B}\Sigma^{\scriptscriptstyle \!-\!1\!/\!2}.
				\end{equation}
				Then, 
				\begin{equation*}
				\begin{aligned}
				\Sigma^{\scriptscriptstyle \!-\!1\!/\!2}\hat{\Sigma}_{\scriptscriptstyle \!B}\Sigma^{\scriptscriptstyle \!-\!1\!/\!2} &= \frac{1}{n_{\scriptscriptstyle \!B}}\sum_{i=1}^{n_{\scriptscriptstyle \!B}} (\Sigma^{\scriptscriptstyle \!-\!1\!/\!2}\mathbf{b}_i)(\Sigma^{\scriptscriptstyle \!-\!1\!/\!2}\mathbf{b}_i)^{\scriptscriptstyle \!\top} \\
				&= \frac{1}{n_{\scriptscriptstyle \!B}}\sum_{i=1}^{n_{\scriptscriptstyle \!B}} \tilde{\mathbf{b}}_i\tilde{\mathbf{b}}_i^{\scriptscriptstyle \!\top} \\
				&= \frac{1}{n_{\scriptscriptstyle \!B}}\tilde{\mathbf{B}}^{\scriptscriptstyle \!\top}\tilde{\mathbf{B}}=\tilde{\Sigma}_{\scriptscriptstyle B},\\
				\end{aligned}
				\end{equation*}
				where $\tilde{\mathbf{b}}_i = \Sigma^{\scriptscriptstyle \!-\!1\!/\!2}\mathbf{b}_i\sim subG(\mathbf{I}_d)$, so $\tilde{\Sigma}_{\scriptscriptstyle B}=\frac{1}{n_{\scriptscriptstyle \!B}}\tilde{\mathbf{B}}^{\scriptscriptstyle \!\top}\tilde{\mathbf{B}}$ is an estimation of $\mathbf{I}_d$. That means that 
				\begin{equation*}
					\begin{aligned}
						&\mathbf{I} \preceq U\Sigma^{\scriptscriptstyle \!-\!1\!/\!2}\hat{\Sigma}_{\scriptscriptstyle \!B}\Sigma^{\scriptscriptstyle \!-\!1\!/\!2}\Longleftrightarrow 1 \leq U\lambda_{min}(\tilde{\Sigma}_{\scriptscriptstyle \!B}), \\
						&L\Sigma^{\scriptscriptstyle \!-\!1\!/\!2}\hat{\Sigma}_{\scriptscriptstyle \!B}\Sigma^{\scriptscriptstyle \!-\!1\!/\!2} \preceq \mathbf{I} \Longleftrightarrow 1 \geq L\lambda_{max}(\tilde{\Sigma}_{\scriptscriptstyle \!B}).
					\end{aligned}
				\end{equation*} 
				From the \textbf{Corollary \ref{cor:secondmoment}}, we know 
				\begin{equation*}
					\begin{aligned}
						&U = \Big(1 - C\Big(\sqrt{\frac{d}{n_{\scriptscriptstyle \!B}}} + \sqrt{\frac{\log(1/\eta)}{n_{\scriptscriptstyle \!B}}}\Big)\Big)^{-2} \Longrightarrow  1 \leq U\lambda_{min}(\tilde{\Sigma}_{\scriptscriptstyle \!B}),\\
						&L= \Big(1 + C\Big(\sqrt{\frac{d}{n_{\scriptscriptstyle \!B}}} + \sqrt{\frac{\log(1/\eta)}{n_{\scriptscriptstyle \!B}}}\Big)\Big)^{-2}\Longrightarrow  1 \geq L\lambda_{max}(\tilde{\Sigma}_{\scriptscriptstyle \!B}),
					\end{aligned}
				\end{equation*}where $C$ is an positive constant only depended by the sub-Gaussian norm.
				$\hfill\square$
			\end{prf}

	\section{The Main Proof}
		\subsection{Proof of Theorem \ref{thm:DPPMTOLSE_main}}

		\begin{prf}
		We prove the three claims separately.

		\paragraph{(i) Privacy.}
		In Algorithm~\ref{alg:DPPMTOLSE}, the public quantities
		\[
		\hat{\Sigma}_{\scriptscriptstyle B}
		=
		\frac{1}{n_{\scriptscriptstyle B}}\mathbf{B}^\top \mathbf{B},
		\qquad
		\hat{\sigma}_{\scriptscriptstyle y,B}
		=
		\left(
		\frac{1}{n_{\scriptscriptstyle B}}
		\|\mathbf{y}_{\scriptscriptstyle \!\!B}\|_2^2
		\right)^{1/2}
		\]
		are computed solely from public data and therefore do not consume privacy budget.

		After the transformed-domain clip, the released private queries are
		\[
		\tilde{\Sigma}_{\scriptscriptstyle A}
		=
		\frac{1}{n_{\scriptscriptstyle A}}
		\tilde{\mathbf A}^{\top}\tilde{\mathbf A},
		\qquad
		\tilde{\Sigma}_{\scriptscriptstyle Ay}
		=
		\frac{1}{n_{\scriptscriptstyle A}}
		\tilde{\mathbf A}^{\top}\tilde{\mathbf y}_{\scriptscriptstyle \!\!A}.
		\]
		By Proposition~\ref{prop:sensitivity_privacy}, their sensitivities satisfy
		\[
		\Delta_{\tilde{\Sigma}_{\scriptscriptstyle A}}
		\le
		\frac{2d\left(1+\log\!\frac{2n_{\scriptscriptstyle A}}{\eta}\right)}
		{n_{\scriptscriptstyle A}},
		\qquad
		\Delta_{\tilde{\Sigma}_{\scriptscriptstyle Ay}}
		\le
		\frac{2\sqrt d\left(1+\log\!\frac{2n_{\scriptscriptstyle A}}{\eta}\right)}
		{n_{\scriptscriptstyle A}}.
		\]
		Hence the Gaussian releases
		\[
		\tilde{\Sigma}_{\scriptscriptstyle A}^{\,DP}
		=
		\tilde{\Sigma}_{\scriptscriptstyle A}+\mathbf G,
		\qquad
		\tilde{\Sigma}_{\scriptscriptstyle Ay}^{\,DP}
		=
		\tilde{\Sigma}_{\scriptscriptstyle Ay}+\mathbf g,
		\]
		with
		\[
		\mathbf G\sim GUE(\sigma_1^2),
		\qquad
		\mathbf g\sim \mathcal N(0,\sigma_2^2 I_d),
		\qquad
		\sigma_1=\frac{\Delta_{\tilde{\Sigma}_{\scriptscriptstyle A}}}{\sqrt{2\rho_1}},
		\qquad
		\sigma_2=\frac{\Delta_{\tilde{\Sigma}_{\scriptscriptstyle Ay}}}{\sqrt{2\rho_2}},
		\]
		satisfy \(\rho_1\)-zCDP and \(\rho_2\)-zCDP respectively. By the composition lemma in Section~\ref{subsec:DP}, their joint release satisfies
		\[
		(\rho_1+\rho_2)\text{-zCDP}
		=
		\rho_{\mathrm{tot}}\text{-zCDP}.
		\]
		This proves part (i).

		\paragraph{(ii) Preservation under clip.}
		We next show that, with high probability, the clip step is inactive.

		For the transformed covariates, Corollary~\ref{cor:dimension_clip} yields that with probability at least \(1-c_{\scriptscriptstyle A}\eta\) with some absolute constant $c_{\scriptscriptstyle A} >0$,
		\[
		\max_{1\le i\le n_{\scriptscriptstyle A}}
		\|\tilde{\mathbf a}_i\|_2
		\le
		r_A(n_{\scriptscriptstyle A},d,\eta).
		\]
		By the definition of the linear model, the response $y_{\scriptscriptstyle A,i}$ is the linear combination of sub-Gaussian covariates and noise. Hence, $y_{\scriptscriptstyle A,i}$ is also a sub-Gaussian variable with a unknown second-moment scalar. For the transformed responses, 
		\[
		\tilde y_{\scriptscriptstyle A,i}
		=
		y_{\scriptscriptstyle A,i}/\hat{\sigma}_{\scriptscriptstyle y,B},
		\]
		together with the sub-Gaussian tail bound and the choice
		\[
		r_y(n_{\scriptscriptstyle A},\eta)
		=
		\sqrt{1+\log\!\frac{2n_{\scriptscriptstyle A}}{\eta}},
		\]
		Corollary~\ref{cor:dimension_clip} in the one-dimensional case yields that there exists an absolute constant \(c_y>0\) such that, with probability at least \(1-c_y\eta\),
		\[
		\max_{1\le i\le n_{\scriptscriptstyle A}}
		|\tilde y_{\scriptscriptstyle A,i}|
		\le
		r_y(n_{\scriptscriptstyle A},\eta).
		\]
		By a union bound, there exists \(c_0>0\) such that with probability at least \(1-c_0\eta\), neither the transformed covariates nor the transformed responses are modified by clip. On this event,
		\[
		\tilde{\Sigma}_{\scriptscriptstyle A}
		=
		\frac{1}{n_{\scriptscriptstyle A}}
		\tilde{\mathbf A}^{\top}\tilde{\mathbf A},
		\qquad
		\tilde{\Sigma}_{\scriptscriptstyle Ay}
		=
		\frac{1}{n_{\scriptscriptstyle A}}
		\tilde{\mathbf A}^{\top}\tilde{\mathbf y}_{\scriptscriptstyle \!\!A},
		\]
		are the non-clipped transformed sufficient statistics, and therefore
		\[
		\tilde{\bm{\beta}}_{\scriptscriptstyle A}
		=
		\tilde{\Sigma}_{\scriptscriptstyle A}^{-1}
		\tilde{\Sigma}_{\scriptscriptstyle Ay}
		\]
		is exactly the transformed-domain OLS estimator. This proves part (ii).

		\paragraph{(iii) Utility under inverse stability.}
		We work on the event where the following three facts hold simultaneously:

		\begin{enumerate}
			\item[(a)] the clip step is inactive, as established in part (ii);
			\item[(b)] the transformed second moment is well-conditioned:
			\[
			\lambda_{\min}(\tilde{\Sigma}_{\scriptscriptstyle A})
			\ge
			L(1-\delta_{A,\eta})^2; 
			\]
			\item[(c)] the privacy noises satisfy the high-probability bounds in Corollary~\ref{cor:noise_bounds}.
		\end{enumerate}

		By Lemma~\ref{lem:precondition_spectrum}, the population transformed covariance is spectrally bounded below by \(L I_d\). Applying the sample second-moment concentration bound from Lemma \ref{lem:singular_Bound} in Appendix~\ref{app:technical_tools} to the transformed design gives, with probability at least \(1-O(\eta)\),
		\[
		\lambda_{\min}(\tilde{\Sigma}_{\scriptscriptstyle A})
		\ge
		L(1-\delta_{A,\eta})^2,
		\]where $\delta_{A,\eta} = C\!\left(
		\sqrt{\frac{d}{n_{\scriptscriptstyle A}}}
		+
		\sqrt{\frac{\log(1/\eta)}{n_{\scriptscriptstyle A}}}
		\right)$ with a fixed positive constant $C$,
		hence
		\begin{equation}\label{eq:app_bound_inv_SigmaAtilde}
		\|\tilde{\Sigma}_{\scriptscriptstyle A}^{-1}\|_2
		\le
		\frac{1}{L(1-\delta_{A,\eta})^2}.
		\end{equation}

		Now decompose the transformed-domain estimation error as
		\[
		\tilde{\bm{\beta}}_{\scriptscriptstyle A}^{\,DP}
		-
		\tilde{\bm{\beta}}_{\scriptscriptstyle A}
		=
		(\tilde{\Sigma}_{\scriptscriptstyle A}+\mathbf G)^{-1}
		(\tilde{\Sigma}_{\scriptscriptstyle Ay}+\mathbf g)
		-
		\tilde{\Sigma}_{\scriptscriptstyle A}^{-1}\tilde{\Sigma}_{\scriptscriptstyle Ay},
		\]
		so that
		\begin{equation}\label{eq:app_error_decomp}
		\begin{aligned}
		\bigl\|
		\tilde{\bm{\beta}}_{\scriptscriptstyle A}^{\,DP}
		-
		\tilde{\bm{\beta}}_{\scriptscriptstyle A}
		\bigr\|_2
		&\le
		\bigl\|
		(\tilde{\Sigma}_{\scriptscriptstyle A}+\mathbf G)^{-1}
		-
		\tilde{\Sigma}_{\scriptscriptstyle A}^{-1}
		\bigr\|_2
		\,
		\|\tilde{\Sigma}_{\scriptscriptstyle Ay}\|_2 \\
		&\qquad
		+
		\|
		(\tilde{\Sigma}_{\scriptscriptstyle A}+\mathbf G)^{-1}
		\|_2
		\,
		\|\mathbf g\|_2.
		\end{aligned}
		\end{equation}

		We first control the matrix-inverse perturbation term. By Corollary~\ref{cor:noise_bounds},
		\[
		\|\mathbf G\|_2
		\le
		C_1
		\frac{d^{3/2}\left(1+\log\!\frac{2n_{\scriptscriptstyle A}}{\eta}\right)}
		{\sqrt{\rho_1}\,n_{\scriptscriptstyle A}}
		=
		C_1
		\frac{d^{3/2}\Gamma_{A,\eta}}
		{\sqrt{\rho_1}\,n_{\scriptscriptstyle A}},
		\]
		up to a change of constants. Together with \eqref{eq:app_bound_inv_SigmaAtilde}, the stability condition \eqref{eq:stability_condition_main} implies
		\[
		\|\tilde{\Sigma}_{\scriptscriptstyle A}^{-1}\|_2\,
		\|\mathbf G\|_2
		\le
		\frac12.
		\]
		Therefore, by Lemma~\ref{lem:inversebound},
		\[
		\|(\tilde{\Sigma}_{\scriptscriptstyle A}+\mathbf G)^{-1}\|_2
		\le
		\frac{\|\tilde{\Sigma}_{\scriptscriptstyle A}^{-1}\|_2}
		{1-\|\tilde{\Sigma}_{\scriptscriptstyle A}^{-1}\|_2\|\mathbf G\|_2}
		\le
		2\|\tilde{\Sigma}_{\scriptscriptstyle A}^{-1}\|_2,
		\]
		and
		\[
		\bigl\|
		(\tilde{\Sigma}_{\scriptscriptstyle A}+\mathbf G)^{-1}
		-
		\tilde{\Sigma}_{\scriptscriptstyle A}^{-1}
		\bigr\|_2
		\le
		\frac{\|\tilde{\Sigma}_{\scriptscriptstyle A}^{-1}\|_2^2\|\mathbf G\|_2}
		{1-\|\tilde{\Sigma}_{\scriptscriptstyle A}^{-1}\|_2\|\mathbf G\|_2}
		\le
		2\|\tilde{\Sigma}_{\scriptscriptstyle A}^{-1}\|_2^2\|\mathbf G\|_2.
		\]
		Using \eqref{eq:app_bound_inv_SigmaAtilde} again gives
		\begin{equation}\label{eq:app_inverse_diff_bound}
		\bigl\|
		(\tilde{\Sigma}_{\scriptscriptstyle A}+\mathbf G)^{-1}
		-
		\tilde{\Sigma}_{\scriptscriptstyle A}^{-1}
		\bigr\|_2
		\le
		C_1
		\frac{d^{3/2}\Gamma_{A,\eta}}
		{\sqrt{\rho_1}\,n_{\scriptscriptstyle A}\,
		L^2(1-\delta_{A,\eta})^4}.
		\end{equation}
	
		Next we control \(\|\tilde{\Sigma}_{\scriptscriptstyle Ay}\|_2\). Recall that
		\[
		\tilde{\Sigma}_{\scriptscriptstyle Ay}
		=
		\hat{\sigma}_{\scriptscriptstyle y,B}^{-1}
		\hat{\Sigma}_{\scriptscriptstyle B}^{-1/2}
		\hat{\Sigma}_{\scriptscriptstyle Ay},
		\]
		so
		\[
		\|\tilde{\Sigma}_{\scriptscriptstyle Ay}\|_2
		\le
		\hat{\sigma}_{\scriptscriptstyle y,B}^{-1}
		\|\hat{\Sigma}_{\scriptscriptstyle B}^{-1/2}\|_2
		\,
		\|\hat{\Sigma}_{\scriptscriptstyle Ay}\|_2.
		\]

		Under the linear model
		\[
		y_i=\mathbf{x}_i^\top\bm{\beta}^\star+\varepsilon_i,
		\]
		we may write
		\[
		\hat{\Sigma}_{\scriptscriptstyle Ay}
		=
		\frac{1}{n_{\scriptscriptstyle A}}\mathbf{A}^\top\mathbf{y}_{\scriptscriptstyle \!\!A}
		=
		\frac{1}{n_{\scriptscriptstyle A}}\mathbf{A}^\top(\mathbf{A}\bm{\beta}^\star+\bm{\varepsilon})
		=
		\hat{\Sigma}_{\scriptscriptstyle A}\bm{\beta}^\star
		+
		\frac{1}{n_{\scriptscriptstyle A}}\mathbf{A}^\top\bm{\varepsilon},
		\]
		where \(\bm{\varepsilon}=(\varepsilon_1,\ldots,\varepsilon_{n_{\scriptscriptstyle A}})^\top\). Hence
		\begin{equation}\label{eq:app_crossmoment_decomp}
		\|\hat{\Sigma}_{\scriptscriptstyle Ay}\|_2
		\le
		\|\hat{\Sigma}_{\scriptscriptstyle A}\|_2\|\bm{\beta}^\star\|_2
		+
		\left\|
		\frac{1}{n_{\scriptscriptstyle A}}\mathbf{A}^\top\bm{\varepsilon}
		\right\|_2.
		\end{equation}

		We first bound \(\|\hat{\Sigma}_{\scriptscriptstyle A}\|_2\) in terms of the population second moment \(\Sigma\). Let
		\[
		\bar{\mathbf A}
		=
		\mathbf{A}\Sigma^{-1/2}.
		\]
		Then the rows of \(\bar{\mathbf A}\) are isotropic sub-Gaussian, and
		\[
		\hat{\Sigma}_{\scriptscriptstyle A}
		=
		\Sigma^{1/2}
		\left(
		\frac{1}{n_{\scriptscriptstyle A}}
		\bar{\mathbf A}^{\top}\bar{\mathbf A}
		\right)
		\Sigma^{1/2}.
		\]
		Therefore, by Corollary~\ref{cor:secondmoment}, with probability at least \(1-2\eta\),
		\begin{equation}\label{eq:app_sigmaA_operator_bound}
		\|\hat{\Sigma}_{\scriptscriptstyle A}\|_2
		\le
		\|\Sigma\|_2
		\left(1+\delta_{A,\eta}\right)^2,
		\end{equation}
		where $\delta_{A,\eta}
		=
		C\left(
		\sqrt{\frac{d}{n_{\scriptscriptstyle A}}}
		+
		\sqrt{\frac{\log(1/\eta)}{n_{\scriptscriptstyle A}}}
		\right)$.

		For the noise term in \eqref{eq:app_crossmoment_decomp}, under the sub-Gaussian design and noise assumptions,
		\[
		\left\|
		\frac{1}{n_{\scriptscriptstyle A}}\mathbf{A}^\top\bm{\varepsilon}
		\right\|_2
		\]
		is of lower order than the leading term
		\(\|\hat{\Sigma}_{\scriptscriptstyle A}\|_2\|\bm{\beta}^\star\|_2\) with high probability, and may therefore be absorbed into the constant. Consequently, with probability at least \(1-O(\eta)\),
		\begin{equation}\label{eq:app_cross_moment_bound}
		\|\hat{\Sigma}_{\scriptscriptstyle Ay}\|_2
		\le
		C'\,
		\|\Sigma\|_2
		\left(1+\delta_{A,\eta}\right)^2
		\|\bm{\beta}^\star\|_2,
		\end{equation}
		for some constant \(C'>0\) depending only on the sub-Gaussian norms.

		Combining \eqref{eq:app_inverse_diff_bound} and \eqref{eq:app_cross_moment_bound}, the first term in \eqref{eq:app_error_decomp} is bounded by
		\begin{equation}\label{eq:app_term1_bound}
		\begin{aligned}
		&
		\bigl\|
		(\tilde{\Sigma}_{\scriptscriptstyle A}+\mathbf G)^{-1}
		-
		\tilde{\Sigma}_{\scriptscriptstyle A}^{-1}
		\bigr\|_2
		\,
		\|\tilde{\Sigma}_{\scriptscriptstyle Ay}\|_2 \\
		&\qquad\le
		C_1
		\frac{d^{3/2}\Gamma_{A,\eta}(1+\delta_{A,\eta})^2}
		{\hat{\sigma}_{\scriptscriptstyle y,B}\,
		\sqrt{\rho_1}\,
		n_{\scriptscriptstyle A}\,
		L^2(1-\delta_{A,\eta})^4}
		\,
		\|\bm{\beta}^\star\|_2
		\|\Sigma\|_2
		\|\hat{\Sigma}_{\scriptscriptstyle B}^{-1/2}\|_2.
		\end{aligned}
		\end{equation}

		For the second term in \eqref{eq:app_error_decomp}, Corollary~\ref{cor:noise_bounds} gives
		\[
		\|\mathbf g\|_2
		\le
		C_2
		\frac{d\left(1+\log\!\frac{2n_{\scriptscriptstyle A}}{\eta}\right)}
		{\sqrt{\rho_2}\,n_{\scriptscriptstyle A}},
		\]
		and therefore
		\begin{equation}\label{eq:app_term2_bound}
		\|(\tilde{\Sigma}_{\scriptscriptstyle A}+\mathbf G)^{-1}\|_2\|\mathbf g\|_2
		\le
		C_2
		\frac{d\left(1+\log\!\frac{2n_{\scriptscriptstyle A}}{\eta}\right)}
		{\sqrt{\rho_2}\,n_{\scriptscriptstyle A}\,
		L(1-\delta_{A,\eta})^2}.
		\end{equation}
		Under any fixed budget split and $d > 1$, and in particular under \(\rho_1=\rho_2=\rho_{\mathrm{tot}}/2\), the term in \eqref{eq:app_term2_bound} is dominated by $\frac{d^{3/2}\Gamma_{A,\eta}(1+\delta_{A,\eta})^2}{\sqrt{\rho_1}n_A L^2(1-\delta)^4}$ up to constants whenever $\Gamma_{A,\eta}$ contains the $\log(1/\eta)$ factor and $\frac{(1+\delta_{A,\eta})}{L(1-\delta_{A,\eta})^2}\ge 1$. Hence it can be absorbed into the constant in the dominant bound. Substituting \eqref{eq:app_term1_bound} and \eqref{eq:app_term2_bound} into \eqref{eq:app_error_decomp}, we obtain
		\[
		\|
		\tilde{\bm{\beta}}_{\scriptscriptstyle A}^{\,DP}
		-
		\tilde{\bm{\beta}}_{\scriptscriptstyle A}
		\|_2
		\le
		C_0
		\frac{
		d^{3/2}\Gamma_{A,\eta}	(1+\delta_{A,\eta})^2
		}{
		\hat{\sigma}_{\scriptscriptstyle y,B}\,
		\sqrt{\rho_1}\,
		n_{\scriptscriptstyle A}\,
		L^2(1-\delta_{A,\eta})^4
		}
		\,
		\|\bm{\beta}^\star\|_2
		\|\Sigma\|_2
		\|\hat{\Sigma}_{\scriptscriptstyle B}^{-1/2}\|_2,
		\]
		which is exactly \eqref{eq:transformed_error_new}.

		Finally, by the post-processing step,
		\[
		\bar{\bm{\beta}}_{\scriptscriptstyle A}^{\,DP}
		=
		\hat{\sigma}_{\scriptscriptstyle y,B}
		\hat{\Sigma}_{\scriptscriptstyle B}^{-1/2}
		\tilde{\bm{\beta}}_{\scriptscriptstyle A}^{\,DP},
		\qquad
		\hat{\bm{\beta}}_{\scriptscriptstyle A}
		=
		\hat{\sigma}_{\scriptscriptstyle y,B}
		\hat{\Sigma}_{\scriptscriptstyle B}^{-1/2}
		\tilde{\bm{\beta}}_{\scriptscriptstyle A}.
		\]
		Hence
		\[
		\bar{\bm{\beta}}_{\scriptscriptstyle A}^{\,DP}
		-
		\hat{\bm{\beta}}_{\scriptscriptstyle A}
		=
		\hat{\sigma}_{\scriptscriptstyle y,B}
		\hat{\Sigma}_{\scriptscriptstyle B}^{-1/2}
		\Bigl(
		\tilde{\bm{\beta}}_{\scriptscriptstyle A}^{\,DP}
		-
		\tilde{\bm{\beta}}_{\scriptscriptstyle A}
		\Bigr),
		\]
		and therefore
		\[
		\|\bar{\bm{\beta}}_{\scriptscriptstyle A}^{\,DP}
		-
		\hat{\bm{\beta}}_{\scriptscriptstyle A}\|_2
		\le
		\hat{\sigma}_{\scriptscriptstyle y,B}
		\|\hat{\Sigma}_{\scriptscriptstyle B}^{-1/2}\|_2
		\,
		\|
		\tilde{\bm{\beta}}_{\scriptscriptstyle A}^{\,DP}
		-
		\tilde{\bm{\beta}}_{\scriptscriptstyle A}
		\|_2.
		\]
		Substituting the transformed-domain bound above yields
		\[
		\|\bar{\bm{\beta}}_{\scriptscriptstyle A}^{\,DP}
		-
		\hat{\bm{\beta}}_{\scriptscriptstyle A}\|_2
		\le
		C_0
		\frac{d^{3/2}}{\sqrt{\rho_1}\,
		n_{\scriptscriptstyle A}\,}
		\frac{
		\Gamma_{A,\eta}(1+\delta_{A,\eta})^2
		}{
		L^2(1-\delta_{A,\eta})^4
		}
		\,
		\|\bm{\beta}^\star\|_2
		\|\Sigma\|_2
		\|\hat{\Sigma}_{\scriptscriptstyle B}^{-1}\|_2,
		\]
		which is exactly \eqref{eq:original_error_new}. This completes the proof.
		\end{prf}
		
	\section{The Baseline Algorithm}\label{app:baseline}
		As a benchmark, we consider the standard private-only sufficient statistics perturbation (SSP) estimator for OLS. The method uses only the private sample to compute empirical scale proxies, clips covariates and responses directly in the original coordinates, forms the private sufficient statistics $\hat{\Sigma}_{\scriptscriptstyle A}=n_{\scriptscriptstyle A}^{\scriptscriptstyle -1}\mathbf{A}^\top \mathbf{A}$ and $\hat{\Sigma}_{\scriptscriptstyle Ay}=n_{\scriptscriptstyle A}^{-1}\mathbf{A}^\top y_{\scriptscriptstyle \!A}$, and then adds Gaussian noise to both quantities under a split privacy budget $\rho_1+\rho_2=\rho_{\mathrm{tot}}$. The final estimator is obtained by inverting the privatized second-moment matrix:
		\[\hat{\beta}_{\scriptscriptstyle A}^{\scriptscriptstyle DP}
		=
		(\hat{\bm \Sigma}_{\scriptscriptstyle A}^{\scriptscriptstyle DP})^{-1}\hat{\Sigma}_{\scriptscriptstyle Ay}^{\scriptscriptstyle DP}.\]
		Its main limitation is that both clip and inverse stability are governed entirely by the private-side geometry. In particular, the clipping radius depend on the scale and trace of $\hat{\Sigma}_{\scriptscriptstyle A}$, and the final error can be strongly amplified when $\hat{\Sigma}_{\scriptscriptstyle A}$ is ill-conditioned.

		\begin{algorithm}[H]
			\small
			\caption{Private-only SSP-OLSE (baseline)}\label{alg:DPOLSE_baseline}
			\begin{algorithmic}[1]
				\STATE {\bfseries Input:} private data $(\mathbf{A},\mathbf{y}_{\scriptscriptstyle \!\!A})$, privacy budget $\rho_{\mathrm{tot}}$, failure probability $\eta$.
				
				\STATE {\bfseries Split privacy budget:}
				choose $\rho_1,\rho_2>0$ such that
				\[
				\rho_1+\rho_2=\rho_{\mathrm{tot}}.
				\]
				For simplicity, one may take $\rho_1=\rho_2=\rho_{\mathrm{tot}}/2$.
				
				\STATE {\bfseries Compute private scale proxies:}
				\[
				\hat{\Sigma}_{\scriptscriptstyle \!A}
				=
				\frac{1}{n_{\scriptscriptstyle \!A}}\mathbf{A}^\top\mathbf{A},
				\qquad
				\hat{\sigma}_{\scriptscriptstyle \!y,A}
				=
				\left(\frac{1}{n_{\scriptscriptstyle \!A}}\|\mathbf{y}_{\scriptscriptstyle \!\!A}\|_2^2\right)^{1/2}.
				\]
				
				\STATE {\bfseries Clipping covariates in the original coordinates:}
				for each row $\mathbf{A}_i$ of $\mathbf{A}$,
				\[
				\mathbf{a}_i
				\leftarrow
				\mathbf{a}_i\cdot
				\min\!\left\{
				1,\,
				\frac{\sqrt{\Tr(\hat{\Sigma}_{\scriptscriptstyle \!A})+d\log(2n_{\scriptscriptstyle \!A}/\eta)}}{\|\mathbf{a}_i\|_2}
				\right\}.
				\]
				
				\STATE {\bfseries Clipping responses in the original coordinates:}
				for each entry $y_{\scriptscriptstyle \!A,i}$ of $\mathbf{y}_{\scriptscriptstyle \!\!A}$,
				\[
				y_{\scriptscriptstyle \!A,i}
				\leftarrow
				y_{\scriptscriptstyle \!A,i}\cdot
				\min\!\left\{
				1,\,
				\frac{\sqrt{\hat{\sigma}_{\scriptscriptstyle \!y,A}^2+\log(2n_{\scriptscriptstyle \!A}/\eta)}}{|y_{\scriptscriptstyle \!A,i}|}
				\right\}.
				\]
				
				\STATE {\bfseries Form private sufficient statistics:}
				\[
				\hat{\Sigma}_{\scriptscriptstyle \!A}
				=
				\frac{1}{n_{\scriptscriptstyle \!A}}\mathbf{A}^\top\mathbf{A},
				\qquad
				\hat{\Sigma}_{\scriptscriptstyle Ay}
				=
				\frac{1}{n_{\scriptscriptstyle \!A}}\mathbf{A}^\top\mathbf{y}_{\scriptscriptstyle \!\!A}.
				\]
				
				\STATE {\bfseries Calibrate Gaussian perturbations:}
				let
				\[
				\Delta_{\hat{\Sigma}_{\scriptscriptstyle \!A}}
				=
				\frac{2\bigl(\Tr(\hat{\Sigma}_{\scriptscriptstyle \!A})+d\log(2n_{\scriptscriptstyle \!A}/\eta)\bigr)}
				{n_{\scriptscriptstyle \!A}},
				\]
				\[
				\Delta_{\hat{\Sigma}_{\scriptscriptstyle Ay}}
				=
				\frac{2\sqrt{\Tr(\hat{\Sigma}_{\scriptscriptstyle \!A})+d\log(2n_{\scriptscriptstyle \!A}/\eta)}
				\sqrt{\hat{\sigma}_{\scriptscriptstyle \!y,A}^2+\log(2n_{\scriptscriptstyle \!A}/\eta)}}
				{n_{\scriptscriptstyle \!A}},
				\]
				and set
				\[
				\sigma_1
				=
				\frac{\Delta_{\hat{\Sigma}_{\scriptscriptstyle \!A}}}{\sqrt{2\rho_1}},
				\qquad
				\sigma_2
				=
				\frac{\Delta_{\hat{\Sigma}_{\scriptscriptstyle Ay}}}{\sqrt{2\rho_2}}.
				\]
				
				\STATE {\bfseries Add Gaussian noise:}
				\[
				\mathbf{G}\sim GUE(\sigma_1^2),
				\qquad
				\mathbf{g}\sim \mathcal N(0,\sigma_2^2\mathbf{I}_d),
				\]
				and define
				\[
				\hat{\Sigma}_{\scriptscriptstyle \!A}^{\,DP}
				=
				\hat{\Sigma}_{\scriptscriptstyle \!A}+\mathbf{G},
				\qquad
				\hat{\Sigma}_{\scriptscriptstyle Ay}^{\,DP}
				=
				\hat{\Sigma}_{\scriptscriptstyle Ay}+\mathbf{g}.
				\]
				
				\STATE {\bfseries Output:}
				\[
				\hat{\bm{\beta}}_{\scriptscriptstyle A}^{\,DP}
				=
				\bigl(\hat{\Sigma}_{\scriptscriptstyle \!A}^{\,DP}\bigr)^{-1}
				\hat{\Sigma}_{\scriptscriptstyle Ay}^{\,DP}.
				\]
			\end{algorithmic}
		\end{algorithm}

\section{Additional comparison with public-DP-gradient baseline}\label{app:dope_compare}

To further contextualize our method relative to recent public-data-assisted DP learning approaches, we include an additional comparison with a DP gradient-based private estimator using the public data by \citet{nasr2023effectively}, named \emph{Differentially Private Origin Estimation-SGD} (DOPE-SGD), but we use the full gradient decent version rather than SGD. Their method uses public data to estimate a clipping origin for privatized gradient updates, thereby reducing the variance of the noisy gradient and improving optimization utility. In contrast, our method uses the public empirical second moment to \emph{precondition the regression geometry} before privatizing sufficient statistics. Thus, the two methods exploit public data in fundamentally different ways: Their method act on the \emph{gradient path}, whereas DP-PMTOLSE acts on the \emph{second-moment geometry}. See also the discussion of public-gradient clipping and public-data-assisted DP optimization in \citet{nasr2023effectively}. The code of all experiments in this paper can be found in \href{https://github.com/Spikelong/Ehence-DP-Linear-Regression.git}{https://github.com/Spikelong/Ehence-DP-Linear-Regression.git}.

\begin{figure}[H]
    \centering
    \includegraphics[width=\textwidth]{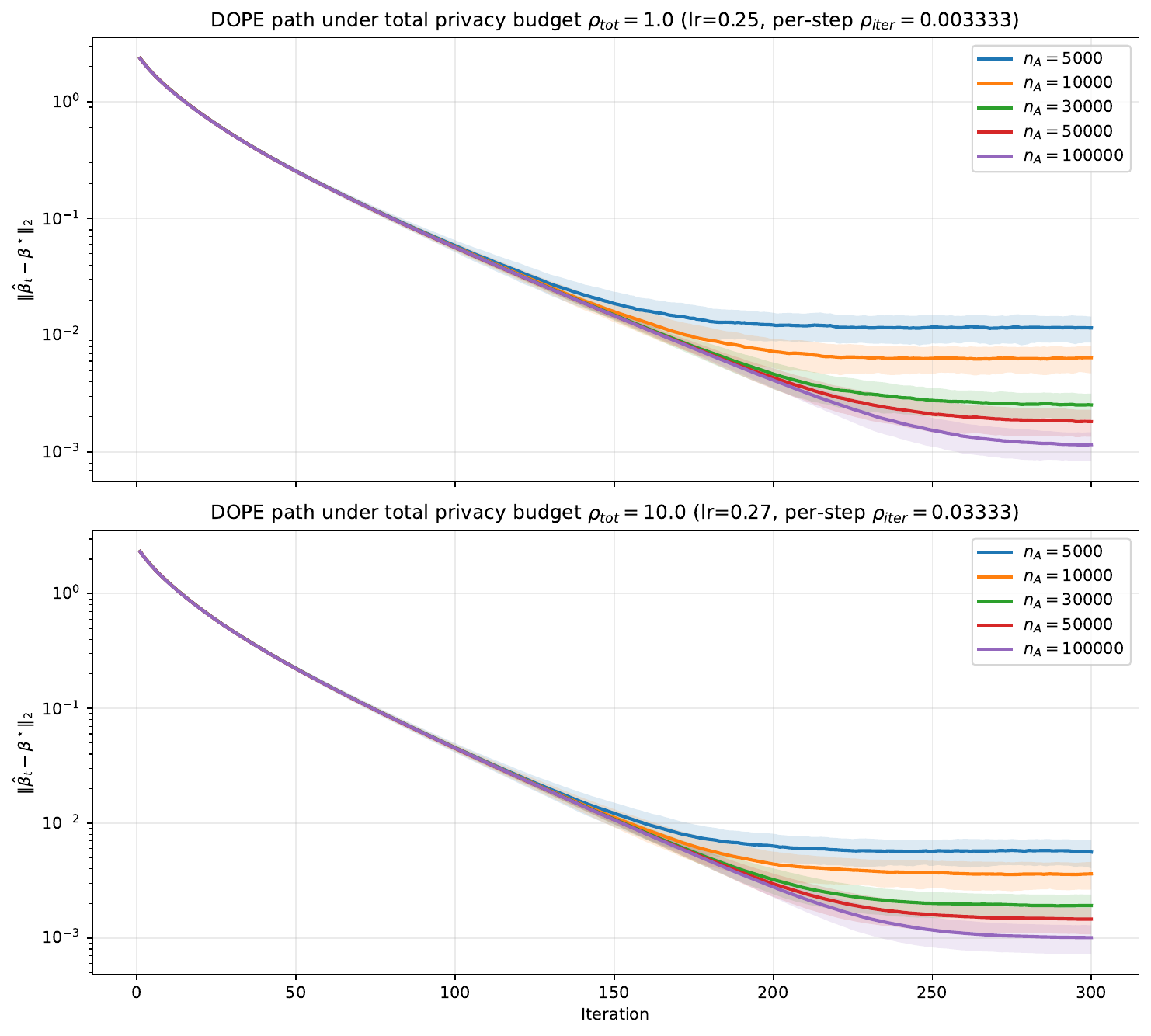}
    \caption{Supplementary iterations of DOPE-style public-gradient baseline in synthetic linear regression. }
    \label{fig:appendix_dope_convergence}
\end{figure}

\textbf{Experimental setup.} We use the same synthetic Gaussian linear model as in Section~\ref{subsec:synthetic_experiment}. The dimension is fixed at \(d=10\), the public sample size is fixed at \(n_B=50\), and the private sample size varies over
\[
n_A \in \{5000,\,10000,\,15000,\,20000,\,25000\}.
\]
The private and public covariates are drawn from the same Gaussian model, and responses are generated from
\[
y = \mathbf{x}^\top \bm \beta^\star + \omega.
\]
We compare the following two methods:
\begin{itemize}
    \item \textbf{DP-PMTOLSE}: our public-second-moment preconditioned SSP estimator;
    \item \textbf{DOPE-GD}: a full-batch private gradient method that clips gradients around a public mean-gradient origin estimated from the public sample.
\end{itemize}
For both methods, the total privacy budget is taken as \(\rho_{\mathrm{tot}} \in \{1,10\}\), and each configuration is repeated over $300$ trials. We tune the DPOE-GD convergence within the $300$ iterations and report some convergence situation in Figure~\ref{fig:appendix_dope_convergence}.

\begin{figure}[H]
    \centering
    \includegraphics[width=\textwidth]{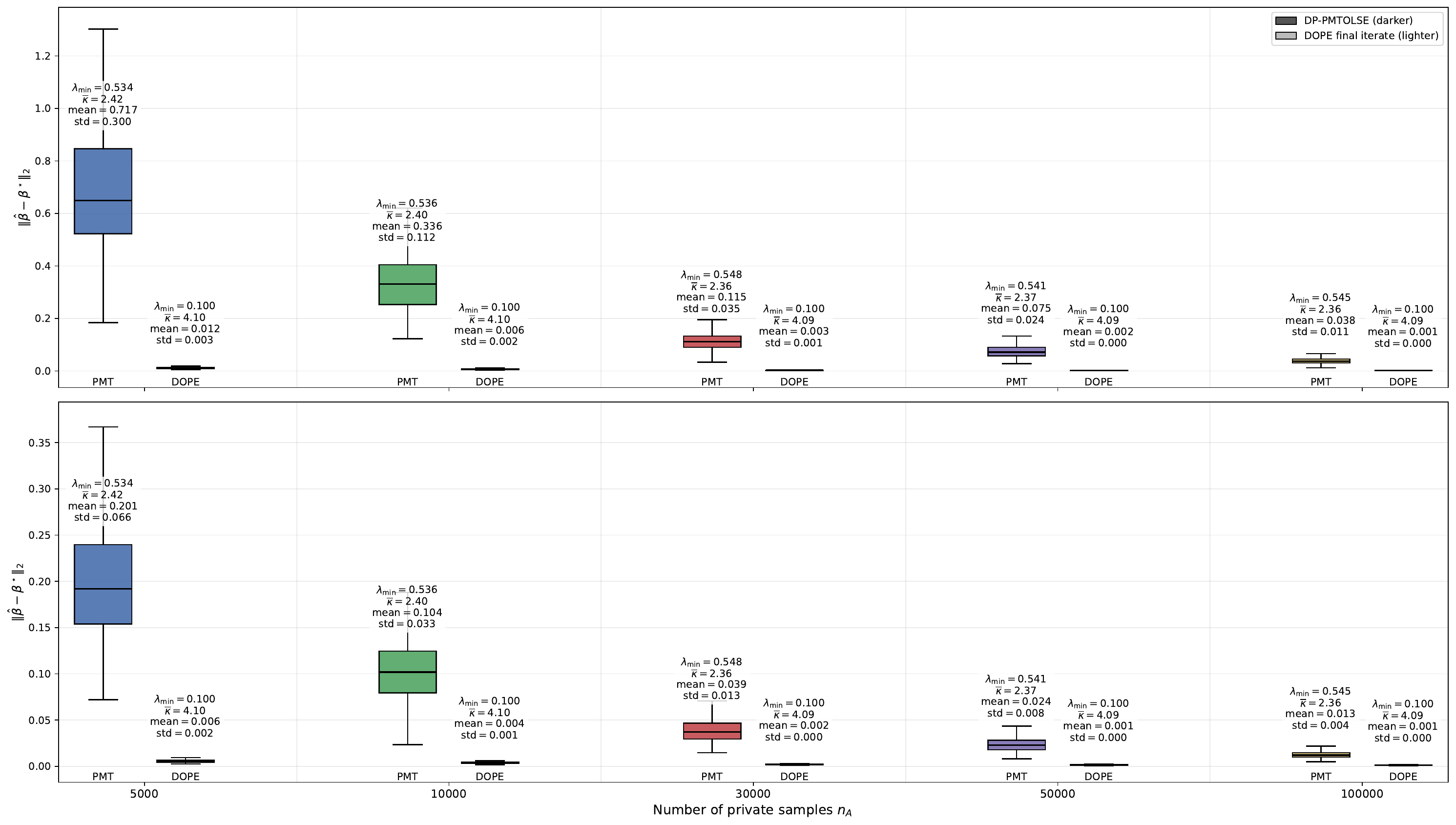}
    \caption{Supplementary estimation comparison between DP-PMTOLSE and a DOPE-style public-gradient baseline in synthetic linear regression. }
    \label{fig:appendix_dope_compare}
\end{figure}

\begin{figure}[H]
    \centering
    \includegraphics[width=\textwidth]{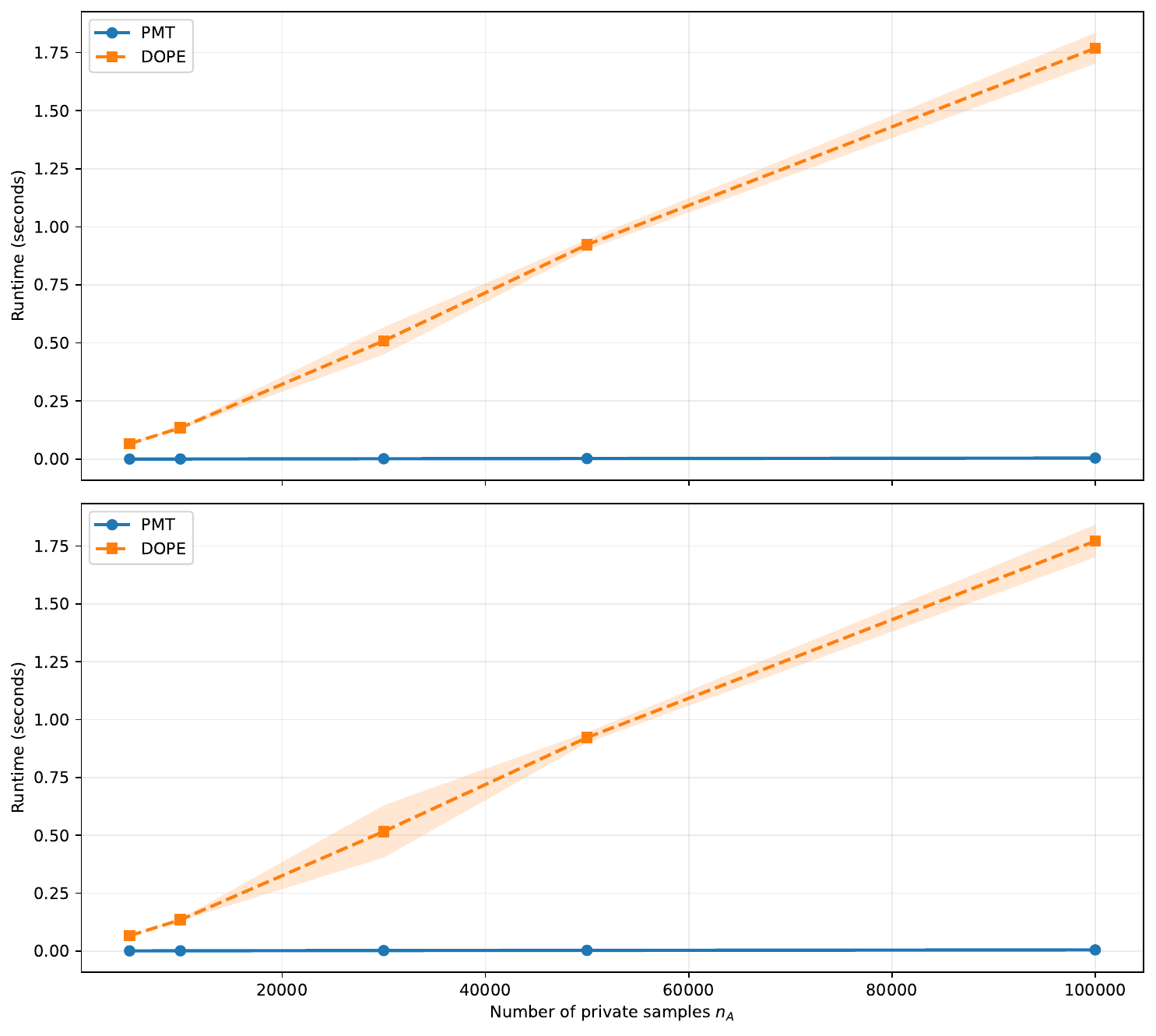}
    \caption{Supplementary runtime comparison between DP-PMTOLSE and a DOPE-style public-gradient baseline in synthetic linear regression. }
    \label{fig:appendix_dope_runtime_compare}
\end{figure}

Figure~\ref{fig:appendix_dope_compare} compares the two methods. The DOPE-style baseline as a DP gradient method produces more accurate and stable results, which is expected because the gradient-based methods have been verified with a better performance in DP situation. However, the parameter-tuning cost and the computation cost are more expensive than SSP-type method. As Figure~\ref{fig:appendix_dope_runtime_compare} showing, the main pattern is that DP-PMTOLSE is substantially faster and typically more stable across all tested private sample sizes. This is expected: DP-PMTOLSE is a closed-form SSP-type estimator that privatizes only transformed sufficient statistics, whereas the DOPE-style baseline requires repeated noisy gradient updates and is therefore computationally more expensive. In addition, the closed-form preconditioned estimator avoids optimization-path variability and benefits directly from improved conditioning of the transformed second-moment matrix.

The comparison highlights a conceptual distinction. The DOPE-style baseline uses public data to improve the \emph{local optimization geometry} of gradient clipping, while DP-PMTOLSE uses public data to improve the \emph{global regression geometry} through second-moment whitening. This supplementary comparison does not claim that DP-PMTOLSE dominates all public-data-assisted DP learning procedures in general. Rather, it shows that for the \emph{specific problem of ordinary least squares under sufficient-statistics perturbation}, public second-moment preconditioning is a more direct and computationally efficient way to use public data than a DOPE-style gradient-based mechanism. The comparison therefore complements the main experiments and further clarifies the methodological distinction between our approach and recent public-gradient DP methods.

\end{appendices}

\bibliography{ref}

\end{document}